\documentclass{article}

\PassOptionsToPackage{numbers, sort&compress}{natbib}
\usepackage[preprint]{neurips_data_2021}

\usepackage[utf8]{inputenc} % allow utf-8 input
\usepackage[T1]{fontenc}    % use 8-bit T1 fonts
\usepackage{url}            % simple URL typesetting
\usepackage{booktabs}       % professional-quality tables
\usepackage{amsfonts}       % blackboard math symbols
\usepackage{amsmath}       % blackboard math symbols
\usepackage{nicefrac}       % compact symbols for 1/2, etc.
\usepackage{microtype}      % microtypography
\usepackage[table]{xcolor}         % colors

\usepackage{soul}           % hl
\usepackage[pdftex]{graphicx}
\usepackage{tabularx}
\usepackage[colorlinks=true,
            linkcolor=black,
            citecolor=black,
            urlcolor=blue,
            ]{hyperref}       % hyperlinks
\usepackage{wrapfig}
\newcolumntype{Z}{>{\raggedright\let\newline\\\arraybackslash\hspace{0pt}}X}
\newcolumntype{L}[1]{>{\raggedright\let\newline\\\arraybackslash\hspace{0pt}}m{#1}}
\usepackage{tcolorbox}
\usepackage[frozencache=true,cachedir=.]{minted}
\tcbuselibrary{minted,breakable,xparse,skins}
\usepackage{amsthm}
\theoremstyle{definition}
\newtheorem{definition}{Definition}[section]

\definecolor{bg}{gray}{0.95}
\usepackage{listings}
\usepackage{setspace} % for \onehalfspacing and \singlespacing macros
\usepackage{etoolbox}
\AtBeginEnvironment{quote}{\singlespacing\small}

\title{Contemporary Symbolic Regression Methods and their Relative Performance}

\author{%
        William La~Cava\thanks{corresponding author} \\
        Institute for Biomedical Informatics\\
        University of Pennsylvania\\
        \texttt{lacava@upenn.edu} \\
        \And
        Patryk Orzechowski \thanks{Department of Automatics and Robotics, AGH University of Science and Technology, Krakow, Poland}\\
        Institute for Biomedical Informatics\\
        University of Pennsylvania\\
        \texttt{patryk.orzechowski@gmail.com} \\
        \And
        Bogdan Burlacu \\
        Josef Ressel Center for Symbolic Regression \\ 
        University of Applied Sciences Upper Austria\\
        \texttt{bogdan.burlacu@fh-ooe.at} \\
        \And
        Fabr\'{i}cio Olivetti de Fran\c{c}a \\
        Federal University of ABC\thanks{Center for Mathematics, Computation and Cognition | Heuristics, Analysis and Learning Laboratory} \\
        Santo Andre, Brazil \\
        \texttt{folivetti@ufabc.edu.br} \\
        \And
        Marco Virgolin \\
        Mechanics and Maritime Sciences \\
        Chalmers University of Technology \\
        \texttt{ marco.virgolin@chalmers.se } \\
        \And
        Ying Jin \\
        Department of Statistics \\
        Stanford University \\ 
        \texttt{ying531@stanford.edu} \\
        \And
        Michael Kommenda  \\
        Josef Ressel Center for Symbolic Regression \\ 
        University of Applied Sciences Upper Austria\\
        \texttt{ michael.kommenda@fh-ooe.at }
        \And
        Jason H. Moore \\ 
        Institute for Biomedical Informatics\\
        University of Pennsylvania \\
        \texttt{ jhmoore@upenn.edu } \\
}

\begin{document}

\maketitle

\begin{abstract}
    Many promising approaches to symbolic regression have been presented in recent years, yet progress in the field continues to suffer from a lack of uniform, robust, and transparent benchmarking standards.
In this paper, we address this shortcoming by introducing an open-source, reproducible benchmarking platform for symbolic regression.
We assess 14 symbolic regression methods and 7 machine learning methods on a set of 252 diverse regression problems. 
Our assessment includes both real-world datasets with no known model form as well as ground-truth benchmark problems, including physics equations and systems of ordinary differential equations. 
For the real-world datasets, we benchmark the ability of each method to learn models with low error and low complexity relative to state-of-the-art machine learning methods. 
For the synthetic problems, we assess each method's ability to find exact solutions in the presence of varying levels of noise. 
Under these controlled experiments, we conclude that the best performing methods for real-world regression combine genetic algorithms with parameter estimation and/or semantic search drivers. 
When tasked with recovering exact equations in the presence of noise, we find that deep learning and genetic algorithm-based approaches perform similarly. 
We provide a detailed guide to reproducing this experiment and contributing new methods, and encourage other researchers to collaborate with us on a common and living symbolic regression benchmark.

\end{abstract}

%TODO

%%%%%%%%%%%%%%%%%%%%%%%%%%%%%%%%%%%%%%%%%%%%%%%%%%%%%%%%%%%%%%%%%%%%%%%%%%%%%%%%
% INSTRUCTIONS
% - please add any additional references to additional.bib, NOT SymbolicRegression.bib
%%%%%%%%%%%%%%%%%%%%%%%%%%%%%%%%%%%%%%%%%%%%%%%%%%%%%%%%%%%%%%%%%%%%%%%%%%%%%%%%
\section{Introduction}
%%%%%%%%%%%%%%%%%%%%%%%%%%%%%%%%%%%%%%%%%%%%%%%%%%%%%%%%%%%%%%%%%%%%%%%%%%%%%%%%

%In regression analysis, SR stands out as an approach that searches for a mathematical expression describing the input-output relationship of the studied phenomena. It differs from most regression models since it does not part from a fixed structure, thus having the possibility of finding a simpler expression.

Symbolic regression (SR) is an approach to machine learning (ML) in which both the parameters and structure of an analytical model are optimized. 
SR can be useful when one wishes to describe a process via a mathematical expression, especially a simple expression; thus, it is often applied in the hopes of producing a model of a process that, by virtue of its simplicity, may be easy to interpret.
Interpretable ML is becoming increasingly important as model deployments in high stakes societal applications such as finance and medicine grow~\cite{jobinGlobalLandscapeAI2019,rudinStopExplaining2019}.  
Moreover, the mathematical expressions produced by SR are well-suited to be analyzed and controlled for their out-of-distribution behavior (e.g., in terms of asymptotic behavior, periodicity, etc.). 
These attractive properties of SR have led to its application in a number of areas, such as physics~\cite{schmidtDistillingFreeformNatural2009}, biology~\cite{schmidtAutomatedModelingStochastic2011}, clinical informatics~\cite{lacavaApplicationConciseMachine2021}, climate modeling~\cite{stanislawskaModelingGlobalTemperature2012a}, finance~\cite{chenGeneticAlgorithmsGenetic2012a}, and many fields of engineering~\cite{smitsParetofrontExploitationSymbolic2005,lacavaAutomaticIdentificationWind2016,castelliFrameworkGeometricSemantic2015}. 

SR literature has, in general, fallen short of evaluating and ranking new methods in a way that facilitates their widespread adoption. 
Our view is that this shortcoming largely stems from a lack of standardized, transparent and reproducible benchmarks, especially those that test a large and diverse array of problems~\cite{mcdermottGeneticProgrammingNeeds2012b}. 
Although community surveys~\cite{whiteBetterGPBenchmarks2012a,mcdermottGeneticProgrammingNeeds2012b} have led to suggestions for improving benchmarking standards, and even black-listed certain problems, contemporary literature continues to be published that violates those standards. %~\cite{petersenDeepSymbolicRegression2020}.
Absent these standards, it is difficult to assess which methods or family of methods should be considered ``state-of-the-art'' (SotA).
% SR literature has, in general, fallen short of evaluating and/or identifying any method or family of methods that could be considered ``state-of-the-art'' (SotA).
% In large part, these shortcomings are due to longstanding issues with the datasets used to benchmark SR, which are often small, trivial, or simply not standardized (i.e. comparable) across papers or experiments~\cite{mcdermottGeneticProgrammingNeeds2012b}.

Achieving a fleeting sense of SotA is certainly not the singular pursuit of methods research, yet without common, robust benchmarking studies, promising avenues of investigation cannot be well-informed by empirical evidence. %the evidence to inform , future research cannot be well-informed. 
We hope the benchmarking platform introduced in this paper improves the cross-pollination between research communities interested in SR, which include evolutionary computation, physics, engineering, statistics, and more traditional machine learning disciplines. 
% Finally, lack of consensus on what SR methods constitute SotA is due in part to the ways in which methodological studies may be conducted - i.e., the relative merit of conducting incremental ablation studies versus large benchmark studies, such as this one. 
% Whatever the causes, the result is that promising directions for future research in SR are not well-informed by empirical evidence. 

%Such datasets do not give an adequate view into the expected performance of the algorithm when applied to datasets whose forms are yet to be discovered. 
In this paper, we describe a large benchmarking effort that includes a dataset repository curated for SR, as well as a benchmarking library designed to allow researchers to easily contribute methods. 
To achieve this, we incorporated 130 datasets with ground truth forms into the Penn Machine Learning Benchmark (PMLB)~\cite{olsonPMLBLargeBenchmark2017d}, including metadata describing the underlying equations, their units, and various summary statistics. 
Furthermore, we created a SR benchmark repository called SRBench\footnote{\url{https://github.com/EpistasisLab/srbench}} and sought contributions from researchers in this area. 
Here we describe this process and the results, which consist of comparisons of 14 contemporary SR methods on hundreds of regression problems. 

To our knowledge, this is by far the largest and most comprehensive SR benchmark effort to date, which allows us to make claims concerning current SotA methods for SR with better certainty. 
Importantly, and in contrast to many previous efforts, the datasets, methods, benchmarking code, and results are completely open-source, reproducible, and revision-controlled, which should allow SRBench to exist as a living benchmark for future studies.

% New methods research risks falling into the trap of poor benchmarking strategies that have plagued previous SR efforts. 

%%%%%%%%%%%%%%%%%%%%%%%%%%%%%%%%%%%%%%%%%%%%%%%%%%%%%%%%%%%%%%%%%%%%%%%%%%%%%%%%
\section{Background and Motivation}
\label{s:background}
%%%%%%%%%%%%%%%%%%%%%%%%%%%%%%%%%%%%%%%%%%%%%%%%%%%%%%%%%%%%%%%%%%%%%%%%%%%%%%%%

The goal of SR is to learn a mapping $\hat{y}(\mathbf{x}) = \hat{\phi}(\mathbf{x}, \hat{\theta}): \mathbb{R}^d \rightarrow \mathbb{R}$ using a dataset of paired examples $\mathcal{D} = \{(\mathbf{x}_i, y_i)\}_{i = 1}^N$, with features $\mathbf{x} \in \mathbb{R}^d$ and target $y$. 
SR assumes the existence of an analytical model of the form $y(\mathbf{x}) = \phi^*(\mathbf{x},\theta^*) + \epsilon$ that would generate the observations in $\mathcal{D}$, and seeks to estimate this model by searching the space of expressions, $\phi$, and parameters, $\theta$, in the presence of white noise, $\epsilon$. 

\citet{kozaGeneticProgrammingProgramming1992a} introduced SR as an application of \textit{genetic programming} (GP), a field that investigates the use of genetic algorithms (GAs) to evolve executable data structures, i.e. programs. 
In the case of so-called ``Koza-style'' GP, the programs to be optimized are syntax trees consisting of functions/operations over input features and constants. 
Like in other GAs, GP is a process that evolves a population of candidate solutions (e.g., syntax trees) by iteratively producing offspring from parent solutions (e.g., by swapping parents' subtrees) and eliminating unfit solutions (e.g., programs with sub-par behavior).
Most SR research to date has emerged from within this sub-field and its associated conferences.\footnote{A non-exhaustive list: \href{https://gecco-2021.sigevo.org/HomePage}{GECCO}, \href{http://www.evostar.org/2021/eurogp/}{EuroGP}, \href{https://dlnext.acm.org/conference/foga}{FOGA}, \href{https://ppsn2020.liacs.leidenuniv.nl/}{PPSN}, and \href{https://cec2021.mini.pw.edu.pl/}{IEEE CEC}.}

Despite the availability of post-hoc methods for explaining black-box model predictions~\cite{lundbergUnifiedApproachInterpreting}, there have been recent calls to focus on learning interpretable/transparent models explicitly~\cite{rudinStopExplaining2019}.
Perhaps due to this renewed interest in model interpretability, entirely different methods for tackling SR have been proposed~\cite{jinBayesianSymbolicRegression2020,petersenDeepSymbolicRegression2020,udrescuAIFeynmanPhysicsInspired2020,panjuAutomatedKnowledgeDiscovery2021,wernerInformedEquationLearning2021,sahooLearningEquationsExtrapolation2018a,kusnerGrammarVariationalAutoencoder2017}.
These include methods based in Bayesian optimization~\cite{jinBayesianSymbolicRegression2020}, recurrent neural networks (RNNs)~\cite{petersenDeepSymbolicRegression2020}, and physics-inspired divide-and-conquer strategies~\cite{udrescuAIFeynmanPhysicsInspired2020,udrescuAIFeynmanParetooptimal2020}. 
Some of these papers refer to Eureqa, a commercial, GP-based SR method used to re-discover known physics equations~\cite{schmidtDistillingFreeformNatural2009}, as the ``gold standard'' for SR~\cite{petersenDeepSymbolicRegression2020} and/or the best method for SR ``by far''~\cite{udrescuAIFeynmanPhysicsInspired2020}.
% One of the most popularized and commercially successful GP-based SR methods is  Eureqa~\cite{schmidtDistillingFreeformNatural2009}, primary used to re-discover known physics equations. This proprietary software acquired by DataRobot in 2017~\footnote{\url{https://www.datarobot.com/nutonian/}}, often referred as the ``gold standard'' for SR~\cite{petersenDeepSymbolicRegression2020} and/or the best method for SR ``by far''~\cite{udrescuAIFeynmanPhysicsInspired2020} has been repetitively shown to be outperformed by the aforementioned works in terms of their ability to discover exact symbolic solutions to synthetic benchmark problems. 
However, \citet{schmidtDistillingFreeformNatural2009b} make no claim to being the SotA method for SR, nor is this hypothesis tested in the body of work on which Eureqa is based~\cite{schmidtMachineScienceAutomated2011}. 

%% Eureqa, what it is and isn't
% On the basis of this claim, the aforementioned works present methods that out-perform Eureqa in terms of their ability to discover exact symbolic solutions to synthetic benchmark problems. 
%However, \citet{schmidtDistillingFreeformNatural2009b} make no claim to being the SotA method for SR, nor is this hypothesis tested in the body of work on which Eureqa is based~\cite{schmidtMachineScienceAutomated2011}.  

Although commercial platforms like Eureqa and Wolfram~\cite{fortunaAutomaticFormulaDiscovery2015} are successful tools for applying SR, they are not designed to support controlled benchmark experiments, and therefore experiments utilizing them have serious caveats.
Due to the design of the front-end API for both tools, it is not possible to benchmark either method against others while holding important parameters of such an experiment constant, including the computational effort, number of model evaluations, CPU/memory limits, and final solution assessment.
More generally, researchers cannot uniquely determine which features of the software and/or experiment lead to observed differences in performance, given that these commercial tools are closed-source.  
In this light, it is not clear what insights are to be gained when comparing to Eureqa and Wolfram beyond a simple head-to-head comparison.
Therefore, rather than benchmark against Eureqa in this paper, we implement its underlying algorithms in an open-source package, which allows our experiment to remain transparent, reproducible, accessible, and controlled.
We discuss the algorithms underlying Eureqa in detail in Sec.~\ref{s:add_background}.

% New methods for SR
A close reading of SR literature since 2009 implies that a number of proposed methods would outperform Eureqa in controlled tests, and are therefore suitable choices for benchmarking (e.g.~\cite{lacavaEpsilonLexicaseSelectionRegression2016c,liskowskiDiscoverySearchObjectives2017}).
% For example, Eureqa's core multi-objective optimization strategy has been outperformed by a number of other optimization methods (e.g.,~\cite{lacavaEpsilonLexicaseSelectionRegression2016c,liskowskiDiscoverySearchObjectives2017}), some of which are included in our study.
% Many other lines of research have pointed in promising directions, including hybridizations of GP with local search for constant optimization~\cite{topchyFasterGeneticProgramming2001,kommendaParameterIdentificationSymbolic2019} or structural tuning~\cite{lacavaInferenceCompactNonlinear2016}; the increasing use of \textit{semantic methods} for guiding variation~\cite{moraglioGeometricSemanticGenetic2012a,virgolinLinearScalingSemantic2019} and selection~\cite{azadKrzysztofKrawiecBehavioral2017,lacavaProbabilisticMultiobjectiveAnalysis2019,arnaldoMultipleRegressionGenetic2014a} (see Section~\ref{s:methods}); and alternative representations of equations~\cite{defrancaInteractionTransformationEvolutionaryAlgorithm2020,mcconaghyFFXFastScalable2011,lacavaLearningConciseRepresentations2019c}.
% The term \textit{semantic methods} refers to evolutionary strategies that leverage the full behavior (i.e. semantics) of the model, and potentially its subcomponents, to drive search, rather than relying on aggregate, scalar fitness values~\cite{vanneschiSurveySemanticMethods2014}.
% lack of standards in SR field
Unfortunately, the widespread adoption of these promising SR approaches is hamstrung by a lack of consensus on good benchmark problems, testing frameworks, and experimental designs. 
Our effort to establish a common benchmark is motivated by our view that common, robust, standardized benchmarks for SR could speed progress in the field by providing a clear baseline from which to assert the quality of new approaches. 
% in the SR community slows down the progress of this field because there is no clear baseline from which to assert the quality of new approaches.
% By contrast, 
% Advances in the NN community that create a ratcheting effect of progress towards better methods and architectures. 
Consider the NN community's focus on common benchmarks (e.g. ImageNet~\cite{dengImagenetLargescaleHierarchical2009}), frameworks (e.g. TensorFlow, PyTorch) and experiment designs. 
By contrast, it is common to observe results in SR literature that are based on a small number of low dimensional, easy and unrealistic problems, comparing only to very basic GP systems such as those described in~\cite{kozaGeneticProgrammingProgramming1992a} nearly thirty years ago.   
Despite detailed descriptions of these issues~\cite{mcdermottGeneticProgrammingNeeds2012b}, community surveys and proposals to ``black-list" toy problems~\cite{whiteBetterGPBenchmarks2012a}, toy datasets and comparisons to out-dated SR methods continue to appear in contemporary literature.
% The issues around benchmarking have been described in detail~\cite{mcdermottGeneticProgrammingNeeds2012b}, and even led to community surveys and a proposal to ``black-list'' several toy problems often reported in literature, such as the quartic polynomial problem~\cite{whiteBetterGPBenchmarks2012a}.

The aspects of performance assessment for SR differ from typical regression benchmarking due to the interest in obtaining concise, symbolic expressions. 
In general, the trade-off between accuracy and simplicity must be considered when evaluating the merits of different models.  
Furthermore, model \emph{simplicity}, typically measured as sparsity or model size, is but a proxy for model \emph{interpretability}; a simple model may still be un-interpretable, or simply wrong~\cite{liptonMythos2018,poursabziManipulating2021,virgolinModelLearning2021}. 
With these concerns in mind, datasets with ground truth solutions are useful, in that they allow researchers to assess whether or not the symbolic model regressed by a given method corresponds to a known analytical solution. 
% Assessing whether the symbolic model returned by a model is an exact match is non-trivial, and requires symbolic solvers equipped with a combination of algebraic simplification and structural pruning techniques.
% In addition, relative fidelity of estimated model forms that are not exact matches are difficult to assess.
Nevertheless, benchmarks utilizing synthetic datasets with ground-truth solutions are not sufficient for assessing real-world performance, and so we consider it essential to also evaluate the performance of SR on real-world or otherwise black-box regression problems, relative to SotA ML methods. 

% efforts to benchmark SR
There have been a few recent efforts to benchmark SR algorithms~\cite{zegklitzBenchmarkingStateoftheartSymbolic2020}, including a precursor to this work benchmarking four SR methods on 94 regression problems~\cite{orzechowskiWhereAreWe2018}. 
In both cases, SR methods were assessed solely on their ability to make accurate predictions. 
In contrast, \citet{udrescuAIFeynmanPhysicsInspired2020} proposed 120 new synthetic, physics-based datasets for SR, but compared only to Eureqa and only in terms of solution rates. 
A major contribution of our work is its significantly more comprehensive scope than previous studies.
We include 14 SR methods on 252 datasets in comparison to 7 ML methods. 
Our metrics of comparison are also more comprehensive, and include 1) accuracy, 2) simplicity, and 3) exact or approximate symbolic matches to the ground truth process. 
Furthermore, we have made the benchmark openly available, reproducible, and open for contributions supported by continuous integration~\cite{fowlerContinuousIntegration2006}. 

%%%%%%%%%%%%%%%%%%%%%%%%%%%%%%%%%%%%%%%%%%%%%%%%%%%%%%%%%%%%%%%%%%%%%%%%%%%%%%%%
\section{SRBench}
\label{s:methods}
%%%%%%%%%%%%%%%%%%%%%%%%%%%%%%%%%%%%%%%%%%%%%%%%%%%%%%%%%%%%%%%%%%%%%%%%%%%%%%%%

We created SRBench to be a reproducible, open-source benchmarking project by pulling together a large set of diverse benchmark datasets, contemporary SR methods, and ML methods around a shared model evaluation and analysis environment.
SRBench overcomes several of the issues in current benchmarking literature as described in Sec.~\ref{s:background}. 
For example, it makes it easy for methodologists to benchmark new algorithms over hundreds of problems, in comparison to strong, contemporary reference methods. 
These improvements allow us to reason with more certainty than in previous work about the SotA methods for SR.

% improve datasets
In order to establish common datasets, we extended PMLB, a repository of standardized regression and classification problems~\cite{olsonPMLBLargeBenchmark2017d,romanoPMLBV1Open2021}, by adding 130 SR datasets with known model forms. 
PMLB provides utilities for fetching and handling data, recording and visualizing dataset metadata, and contributing new datasets. 
% improve SR software 
The SR methods we benchmarked are all contemporary implementations (2011 - 2020) from several method families, as shown in Tbl.~\ref{tbl:methods}. 
We required contributors to implement a minimal, Scikit-learn compatible~\cite{pedregosaScikitlearnMachineLearning2011a}, Python API for their method.
In addition, contributors were required to provide the final fitted model as a string that was compatible with the symbolic mathematics library \href{https://www.sympy.org/en/index.html}{sympy}.  
Note that although we require a Python wrapper, SR implementations in many different languages are supported, as long as the Python API is available and the language environment can be managed via Anaconda\footnote{https://www.anaconda.com/}.

% table of methods
\begin{table}
    % \footnotesize
    \scriptsize
    \center
    \caption{
        Short descriptions of the SR methods benchmarked in our experiment, including references and links to implementations. 
    }\label{tbl:methods}
    \rowcolors{2}{gray!25}{white}
    \begin{tabular}{l l L{18em} l r } %L{12em} L{12em} r}
        \rowcolor{white}
        Method      &   Year    &   Description                                         &   Method Family                       &   Implementation  \\ 
        \midrule
        AFP~\cite{schmidtAgefitnessParetoOptimization2011}               &  2011    &   Age-fitness Pareto Optimization     &   GP
                    &   C++/Python (\href{https://github.com/EpistasisLab/ellyn}{link})                  \\
        AFP\_FE~\cite{schmidtDistillingFreeformNatural2009b}             &  2011 &   AFP with co-evolved fitness estimates; Eureqa-esque     &   GP
                    &   C++/Python (\href{https://github.com/EpistasisLab/ellyn}{link})                  \\
        AIFeynman~\cite{udrescuAIFeynmanParetooptimal2020}               &  2020 &   Physics-inspired method             &   Divide and conquer
                    &   Fortran/Python (\href{https://github.com/SJ001/AI-Feynman}{link})              \\
        BSR~\cite{jinBayesianSymbolicRegression2020}                     &  2020 &   Bayesian Symbolic Regression        &   Markov Chain Monte Carlo
                    &   Python (\href{https://github.com/ying531/MCMC-SymReg}{link})                      \\
        DSR~\cite{petersenDeepSymbolicRegression2020}                    &  2020    &   Deep Symbolic Regression            &   Recurrent neural networks   
                    &   Python (PyTorch) (\href{https://github.com/brendenpetersen/deep-symbolic-regression}{link})            \\
        EPLEX~\cite{lacavaProbabilisticMultiobjectiveAnalysis2019}       &  2016    &   $\epsilon$-lexicase selection       &   GP
                    &   C++/Python (\href{https://github.com/EpistasisLab/ellyn}{link})                  \\
        FEAT~\cite{lacavaLearningConciseRepresentations2019}            &   2019    &   Feature Engineering Automation Tool &   GP                          
                    &   C++/Python (\href{https://github.com/lacava/feat}{link})                  \\
        FFX~\cite{mcconaghyFFXFastScalable2011}         &   2011    &   Fast function extraction             &   Random search               
                    &   C++/Python (\href{https://github.com/natekupp/ffx/tree/master/ffx}{link})                  \\
        GP-GOMEA~\cite{virgolin2020improving}     & 2020    &   GP version of the Gene-pool Optimal Mixing Evolutionary Algorithm     & GP            
                    &   C++/Python (\href{https://github.com/marcovirgolin/GP-GOMEA/}{link})                  \\
        gplearn     &   2015    &   Koza-style symbolic regression in Python       &   GP                          
                    &   C++/Python (\href{https://github.com/trevorstephens/gplearn}{link})                  \\
        ITEA~\cite{defrancaInteractionTransformationEvolutionaryAlgorithm2020}   &  2020    &   Interaction-Transformation EA       &   GP
                    &   Haskell/Python (\href{https://github.com/folivetti/ITEA/}{link})              \\
        MRGP~\cite{arnaldoMultipleRegressionGenetic2014a}                        &  2014    &   Multiple Regression Genetic Programming &   GP
                    &   Java (\href{https://github.com/flexgp/gp-learners}{link})                        \\
        Operon~\cite{kommendaParameterIdentificationSymbolic2019}                &  2019    &   SR with Non-linear least squares     &   GP
                    &   C++/Python (\href{https://github.com/heal-research/operon}{link})                  \\
        SBP-GP~\cite{virgolinLinearScalingSemantic2019}                     &   2019    & Semantic Back-propagation Genetic Programming           &   GP
                    &   C++/Python (\href{https://github.com/marcovirgolin/GP-GOMEA}{link})                  \\ 
        \bottomrule
    \end{tabular}
\end{table}

% speed adoption
To ensure reproducibility, we defined a common environment (via Anaconda) with fixed versions of packages and their dependencies. 
In contrast to most SR studies, the full installation code, experiment code, results and analysis are available via the repository for use in future studies. 
% By incorporating diverse approaches to SR in our benchmark, we seek to answer unanswered questions about SotA SR methods while also encouraging and promoting the unification of benchmark practices among researchers who typically work in separate, parallel fields. 
% In order to foster connections between the disparate research communities interested in SR, we have released SRBench with clear documentation and contribution guidelines, and with support for multiple cluster environments and programming languages. 
To make SRBench as extensible as possible, we automated the process of incorporating new methods and results into the analysis pipeline. 
The repository accepts rolling contributions of new methods that meet the minimal API requirements. 
To achieve this, we created a continuous integration (CI)~\cite{fowlerContinuousIntegration2006} framework that assures contributions are compatible with the benchmark code as they arrive.  
CI also supports continuous updates to results reporting and visualization whenever new experiments are available, allowing us to maintain a standing leader-board of contemporary SR methods. 
Ideally these features will quicken the adoption of SotA approaches throughout the SR research community.
Further details on how to use and contribute to SRBench are provided in Sec.~\ref{s:howto}.

% We can thus hope to understand the differences in performance among contemporary SR methods in order to properly inform and motivate future investigations. 

%%%%%%%%%%%%%%%%%%%%%%%%%%%%%%%%%%%%%%%%%%%%%%%%%%%%%%%%%%%%%%%%%%%%%%%%%%%%%%%%
\section{Experiment Design}
%%%%%%%%%%%%%%%%%%%%%%%%%%%%%%%%%%%%%%%%%%%%%%%%%%%%%%%%%%%%%%%%%%%%%%%%%%%%%%%%

We evaluated SR methods on two separate tasks. 
First, we assessed their ability to make accurate predictions on ``black-box'' regression problems (in which the underlying data generating function remains unknown) while minimizing the complexity of the discovered models. 
Second, we tested the ability of each method to find exact solutions to synthetic datasets with known, ground-truth functions, originating from physics and various fields of engineering.
%The second set of problems are based on physics equations from several sources, as described below. 

The basic experiment settings are summarized in Tbl.~\ref{tbl:exp}.
Each algorithm was trained on each dataset (and level of noise, for ground-truth problems) in 10 repeated trials with a different random state that controlled both the train/test split and the seed of the algorithm.  
Datasets were split 75/25\% in training and testing. 
For black-box regression problems, each algorithm was tuned using 5-fold cross validation with halving grid search. 
The SR algorithms were limited to 6 hyperparameter combinations; the ML methods were allowed more, as shown in Tbls.~\ref{tbl:ml_methods}-\ref{tbl:sr_methods2}. 
The best hyperparameter settings were used to tune a final estimator and evaluate it according to the metrics described above. 
Details for running the experiment are given in Sec.~\ref{s:howto}. 

\begin{table}
    \scriptsize
    \center
    \caption{
        Settings used in the benchmark experiments. 
        ``Total comparisons" refers to the total evaluatons of an algorithm on a dataset for a given noise level and random seed.
    }\label{tbl:exp}
    % \rowcolors{2}{gray!25}{white}
    \begin{tabular}{lll}
        Setting                     &   Black-box Problems              &   Ground-truth Problems                   \\
        \midrule
        No. of datasets             &   122                             &   130                                     \\
        No. of algorithms           &   21 (14 SR, 7 ML)                &   14                                      \\
        No. of trials per dataset   &   10                              &   10                                      \\
        Train/test Split            &   .75/.25                         &   .75/.25                                 \\
        Hyperparameter Tuning       &   5-fold Halving Grid Search CV   &   Tuned set from Black-box problems       \\
        Termination criteria        &   500k evaluations/train or 48 hours    &   1M evaluations or 8 hours         \\ 
        Levels of target noise      &   None                            &   0, 0.001, 0.01, 0.1                     \\
        Total comparisons           &   26840                           &   54600                                   \\ 
        Computing Budget            &   1.29M core hours                &   436.8K core hours                       \\  
        \bottomrule
    \end{tabular}
\end{table}

\subsection{Symbolic Regression Methods}

Here we characterize the SR methods summarized in Tbl.~\ref{tbl:methods} by describing how they fit into broader research trends within the SR field.   
The most traditional implementation of GP-based SR we test is \textbf{gplearn}, which initializes a random population of programs/models, and then iterates through the steps of tournament selection, mutation and crossover.
% We include \textbf{gplearn} as an implementation of this traditional style of GP in our benchmark. 
% Selection chooses models from which to produce new points in the search space in the subsequent iteration. 
% The de-facto method of selection is tournament selection, in which a fixed-size sample of models is drawn and the fittest model chosen as winner. 
% Selected models then undergo some combination of randomized mutation and/or crossover, traditionally at the subtree level or single node level.
% In Koza-style GP, the resulting offspring replace the parents, but several common methods include a survival step in which the resulting offspring compete against the parents, at the very least to ensure the best model in the search space is not lost.
% However, most contemporary GP-based SR methods differ in multiple ways, which we summarize broadly below. 

% \paragraph{Pareto Optimization}

Pareto optimization methods~\cite{debFastElitistNondominated2000,smitsParetofrontExploitationSymbolic2005,zitzlerSPEA2ImprovingStrength2001,bleulerMultiobjectiveGeneticProgramming2001} are popular evolutionary strategies that exploit Pareto dominance relations to drive the population of models towards a set of efficient trade-offs between competing objectives.
Half of the SR methods we test use Pareto optimization in some form during training. 
% In combination with random restarts, a model's age (the number of generations since its oldest component entered the population) as an objective has also proven to be an effective way to reduce premature convergence, as well as bloat~\cite{hornbyALPSAgelayeredPopulation2006}. 
% In particular, Eureqa demonstrated a significant improvement over traditional tournament-based selection through its introduction of age-fitness Pareto optimization (\textbf{AFP}). 
Age-Fitness Pareto optimization (\textbf{AFP}), proposed by Eureqa's authors~\citet{schmidtAgefitnessParetoOptimization2011}, uses a model's age as an objective in order to reduce premature convergence as well as bloat~\cite{hornbyALPSAgelayeredPopulation2006}.
% A new model is introduced each generation and prevents it from competing against older, more accurate equations by using age as an objective. 
\textbf{AFP\_FE} combines AFP with Eureqa's method for fitness estimation~\cite{schmidtCoevolutionFitnessPredictors2008}.
Thus we expect AFP\_FE and AFP to perform similarly to Eureqa as described in literature. 
% AFP, AFP_FE, AIFEYNMAN, EPLEX, FEAT, FFX, MRGP, others?
% Note that Eureqa, AFP, AFP\_FE, FEAT, and EPLEX also maintain a passive Pareto archive balancing accuracy/complexity during training that is used to select the final model.

% \paragraph{Semantic Search Drivers}

Another promising line of research has been to leverage program \textit{semantics} (in this case, the equation's intermediate and final outputs over training samples) more heavily during optimization, rather than compressing that information into aggregate fitness values~\cite{azadKrzysztofKrawiecBehavioral2017}. 
$\epsilon$-lexicase selection (\textbf{EPLEX})~\cite{lacavaEpsilonLexicaseSelectionRegression2016c} is a parent selection method that utilizes semantics to conduct selection by filtering models through randomized subsets of cases, which rewards models that perform well on difficult regions of the training data. 
EPLEX is also used as the parent selection method in FEAT~\cite{lacavaLearningConciseRepresentations2019}. 
Semantic backpropagation (SBP) is another semantic technique to compute, for a given target value and a tree node position, that value which makes the output of the model match the target (i.e., the label)~\cite{wieloch2013running,krawiec2013approximating,pawlak2014semantic}. 
% It works by chaining inversions of tree nodes from the root to a chosen node position. 
Here, we evaluate the (\textbf{SBP-GP}) algorithm by~\citet{virgolinLinearScalingSemantic2019} which improves SBP-based recombination by dynamically adapting intermediate outputs using affine transformations. 
%a mutation operator that substitutes desired semantics into programs from a pre-computed library of trees. 
% Semantic backpropagation GP (\textbf{SBP}) determines the desirable output for a given tree node by backpropagating it to specific node using inversions of function found along the way ~\cite{wieloch2013running,krawiec2013approximating,pawlak2014semantic}. 
% In this paper, an SBP implementation proposed by Virgolin et al. ~\cite{virgolinLinearScalingSemantic2019} is used, which is an extension of ~\cite{wieloch2013running}. This version uses a pre-computed library of trees for substitution as a variational operator.

% \paragraph{Constant optimization}
%Traditional GP encodes constants as building blocks that are initialized from a given distribution and optimized along other operators.
%, and then optimizes them through the same process as the rest of the component building blocks.
Backpropagation-based gradient descent was proposed for GP-SR by~\citet{topchyFasterGeneticProgramming2001}, but tends to appear less often than stochastic hill climbing (e.g.~ \cite{bongardNonlinearSystemIdentification2005a,schmidtDistillingFreeformNatural2009}).
%For example, the methods of~\citet{bongardNonlinearSystemIdentification2005a} and~\citet{schmidtDistillingFreeformNatural2009} that would appear years later opted for stochastic hill climbing for handling constant optimization. 
More recent studies~\cite{kommendaEffectsConstantOptimization06,kommendaParameterIdentificationSymbolic2019} have made a strong case for the use of gradient-based constant optimization as an improvement over stochastic and evolutionary approaches.
The aforementioned studies are embodied by \textbf{Operon}, a GP method that incorporates non-linear least squares constant optimization using the Levenberg-Marquadt algorithm~\cite{burlacuOperonEfficientGenetic2020}.

% \paragraph{Novel Model Encodings}

In addition to the question of how to best optimize constants, a line of research has proposed different ways of defining program and/or model encodings. 
% Rather than treating constants as building blocks, \textbf{FEAT}~\cite{lacavaLearningConciseRepresentations2019c} and \textbf{Operon}~\cite{burlacuOperonEfficientGenetic2020} encode weights as edges, and differ in their assignment of weights to all nodes (FEAT) versus just input variables (Operon). 
The methods FEAT, MRGP, ITEA, and FFX each impose additional structural assumptions on the models being evolved. 
In \textbf{FEAT}, each model is a linear combination of a set of evolved features, the parameters of which are encoded as edges and optimized via gradient descent.
In \textbf{MRGP}~\cite{arnaldoMultipleRegressionGenetic2014a}, the entire program trace (i.e., each subfunction of the model) is decomposed into features and used to train a Lasso model.
In \textbf{ITEA}, each model is an affine combination of \textit{interaction-transformation} expressions, which compose a unary function (the transformation) and a polynomial function (the interaction)~\cite{defrancaGreedySearchTree2018,defrancaInteractionTransformationEvolutionaryAlgorithm2020}.
Finally, \textbf{FFX}~\cite{mcconaghyFFXFastScalable2011} simply initializes a population of equations, selects the Pareto optimal set, and returns a single linear model by treating the population of equations as features.

% \paragraph{Adaptive Recombination}

\textbf{GP-GOMEA} is a GP algorithm where recombination is adapted over time~\cite{virgolin2017scalable,virgolin2020improving}. 
Every generation, GP-GOMEA builds a statistical model of interdependencies within the encoding of the evolving programs, and then uses this information to recombine interdependent blocks of components, as to preserve their concerted action.
%A linkage model is a model of the level of interdependencies (i.e., linkage) existing within the encoding of evolving solutions~\cite{thierens2011optimal}.
%Every generation, GP-GOMEA updates its linkage model and then uses it to drive the recombination process, mixing blocks of components that appear to be interdependent.
% GP-GOMEA was found to be especially competitive when relatively small programs are sought. %~\cite{virgolin2017scalable,virgolin2020improving}.

% In addition to GP-based and stochastic SR methods, we tested three contemporary SR methods based in other fields of optimization, summarized below. 

% \paragraph{Bayesian Symbolic Regression}

\citet{jinBayesianSymbolicRegression2020} recently proposed Bayesian Symbolic Regression (\textbf{BSR}), in which a prior is placed on tree structures and the posterior distributions are sampled using a Markov Chain Monte Carlo (MCMC) method.   
As in GP-based SR, arithmetic expressions are expressed with symbolic trees, although BSR explicitly defines the final model form as a linear combination of several symbolic trees. 
Model parsimony is encouraged by specifying a prior that presumes additive, linear combinations of small components. 

% \paragraph{Deep Symbolic Regression}
Deep Symbolic Regression (\textbf{DSR})~\cite{petersenDeepSymbolicRegression2020} uses reinforcement learning to train a generative RNN model of symbolic expressions. 
Expressions sampled from the model distribution are assessed to create a reward signal.  
DSR introduces a variant of the Monte Carlo policy gradient algorithm~\cite{williamsSimpleStatisticalGradientfollowing1992} dubbed a ``risk-seeking policy gradient" in an effort to bias the generative model towards exact expressions.    

% \paragraph{AI Feynman}
\textbf{AIFeynman} is a divide-and-conquer approach that recursively applies a set of solvers and problem decomposition heuristics to build a symbolic model~\cite{udrescuAIFeynmanPhysicsInspired2020}. 
If the problem is not directly solve-able by polynomial fitting or brute-force search, AIFeynman trains a NN on the data and uses it to estimate functional modularities (e.g., symmetry and/or separability), which are used to partition the data into simpler problems and recurse. 
An updated version of the algorithm, which we test here, integrates Pareto optimization with an information-theoretic complexity metric to improve robustness to noise~\cite{udrescuAIFeynmanParetooptimal2020}.
% The method applies multiple techniques to symbolically solve the regression problem, including dimensionality reduction, polynomial fit or brute-force approach
% If solution can not be determined, a feed-forward neural network is applied to handle more complex problems~\cite{udrescuAIFeynmanPhysicsInspired2020}.
% An updated AI-Feynman 2.0 version uses Pareto-optimal solutions to balance between the accuracy and complexity of the solutions ~\cite{udrescuAIFeynmanParetooptimal2020}.
\subsection{Datasets}

All of the benchmark datasets are summarized by number of instances and number of features in Fig.~\ref{fig:pmlb}. 
The problems range from 47 to 1 million instances, and two to 124 features.  
% \paragraph{Black-box Regression Problems}
We used 122 black-box regression problems available in PMLB v.1.0. 
These problems are pulled from, and overlap with, various open-source repositories, including OpenML~\cite{vanschorenOpenMLNetworkedScience2013} and the UCI repository~\cite{lichmanUCIMachineLearning2013a}. 
PMLB standardizes these datasets to a common format and provides fetching functions to load them into Python (and R). 
The black-box regression datasets consist of 46 ``real-world" problems (i.e., observational data collected from physical processes) and 76
synthetic problems (i.e., data generated computationally from static functions or simulations).
The black-box problems cover diverse domains, including health informatics (11), business (10), technology (10), environmental science (11) and government (12); in addition, they are derived from varied data sources, including human subjects (14), environmental observations (11), government studies (12), and economic markets (7).
The datasets can be browsed by their properties at \href{https://epistasislab.github.io/pmlb}{epistasislab.github.io/pmlb}.
Each dataset includes metadata describing source information as well as a detailed profile page summarizing the data distributions and interactions (\href{https://epistasislab.github.io/pmlb/profile/analcatdata_aids.html}{here is an example}).

% \paragraph{Ground-Truth Regression Problems}
We extended PMLB with 130 datasets with known, ground-truth model forms. 
These datasets were used to assess the ability of SR methods to recover known process physics. 
The 130 datasets came from two sources: the \href{https://space.mit.edu/home/tegmark/aifeynman.html}{Feynman Symbolic Regression Database}, 
and the \href{https://github.com/lacava/ode-strogatz}{ODE-Strogatz repository}.
Both sets of data come from first principles models of physical systems. 
The Feynman problems originate in the \textit{Feynman Lectures on Physics}~\cite{feynmanFeynmanLecturesPhysics2015}, and the datasets were recently created and proposed as SR benchmarks~\cite{udrescuAIFeynmanPhysicsInspired2020}. 
Whereas the Feynman datasets represent static systems, the Strogatz problems are non-linear and chaotic dynamical processes~\cite{strogatzNonlinearDynamicsChaos2014}.
Each dataset is one state of a 2-state system of first-order, ordinary differential equations (ODEs). 
They were used to benchmark SR methods in previous work~\cite{lacavaInferenceCompactNonlinear2016,schmidtMachineScienceAutomated2011}, and are described in more detail in Sec.~\ref{s:add_dataset} and Tbl.~\ref{tbl:strogatz}.

\subsection{Metrics}
\paragraph{Accuracy}
We assessed accuracy using the coefficient of determination, defined as 
% $    R^2 = 1 - \frac{\sum_i^N{(y_i - \hat{y}_i)^2}}{\sum_i^N{(y_i - \bar{y}_i)^2}}.$
% \begin{equation}
% \label{eq:r2}
\[
    R^2 = 1 - \frac{\sum_i^N{(y_i - \hat{y}_i)^2}}
                   {\sum_i^N{(y_i - \bar{y}_i)^2}}.
\]
% \end{equation}

%%
% complexity
%%
\paragraph{Complexity}
A number of different complexity measures have been proposed for SR, including those based on \textit{syntactic} complexity (i.e. related to the complexity of the symbolic model); those based on \textit{semantic} complexity (i.e. related to the behavior of the model over the data)~\cite{vladislavlevaOrderNonlinearityComplexity2009a,udrescuAIFeynmanParetooptimal2020}; those using both definitions~\cite{kommendamichaelEvolvingSimpleSymbolic2015}; and those estimating complexity via meta-learning~\cite{virgolinLearningAFormula2020}. 
The pros and cons of these methods and their relation to notions of interpretability is a point of discussion~\cite{murdochDefinitionsMethodsApplications2019}. 
For the sake of simplicity, we opted to define complexity as the number of mathematical operators, features and constants in the model, where the mathematical operators are in the set 
$\{+,
    -,
    *,
    {/},
    \sin,
    \cos,
    \arcsin,
    \arccos,
    \exp,
    \log, 
\text{pow},
\max,
\min \}$. 
In addition to calculating the complexity of the raw model forms returned by each method, we calculated the complexity of the models after simplifying via sympy. %\href{https://www.sympy.org/en/index.html}{sympy}.
% As mentioned earlier, this simplification procedure is approximate and not always successful; see the Appendix for further details. 

%%
% solutions
%%%
\paragraph{Solution Criteria}
For the ground-truth regression problems, we used the following solution definition.

\begin{definition}[Symbolic Solution]\label{def:soln}
    A model $\hat{\phi}(\mathbf{x}, \hat{\theta})$ is a Symbolic Solution to a problem with ground-truth model $y = \phi^*(\mathbf{x}, \theta^*) + \epsilon$, if $\hat{\phi}$ does not reduce to a constant, and if either of the following conditions are true: 1) $\phi^*-\hat{\phi} = a $; or 2) $\phi^*/\hat{\phi} = b, b \neq 0$, for some constants $a$ and $b$.
\end{definition}

This definition is designed to capture models that differ from the true model by a constant or scalar. 
Prior to assessing symbolic solutions, each model underwent sympy simplification, as did the conditions above. 
% We chose to report both solution rates and test set $R^2$ for these problems because there are pros and cons to both. 
Relative to accuracy metrics, the Symbolic Solution metric is a more faithful evaluation of the ability of an SR method to discover the data generating process.
However, because models can be represented in myriad ways, and sympy's simplification procedure is non-optimal, we cannot guarantee that all symbolic solutions are captured with perfect fidelity by this metric. 
% The accuracy metric more explicitly measures the model's ability to capture the underlying process, yet it has the disadvantage of being reported over finite samples. 
% Therefore accuracy may not truly capture the data generating process. 
% Thus, one may view the accuracy and symbolic solution metrics as optimistic and pessimistic estimates of performance, respectively.
%%%%%%%%%%%%%%%%%%%%%%%%%%%%%%%%%%%%%%%%%%%%%%%%%%%%%%%%%%%%%%%%%%%%%%%%%%%%%%%%
\section{Results}
%%%%%%%%%%%%%%%%%%%%%%%%%%%%%%%%%%%%%%%%%%%%%%%%%%%%%%%%%%%%%%%%%%%%%%%%%%%%%%%%
%%%%%%%%%%%%%%%%%%%%%%%%%%% RESULTS FIGS %%%%%%%%%%%%%%%%%%%%%%%%%%%%%%%%%%%%%%%
\begin{figure}
    \includegraphics[width=0.9\textwidth]{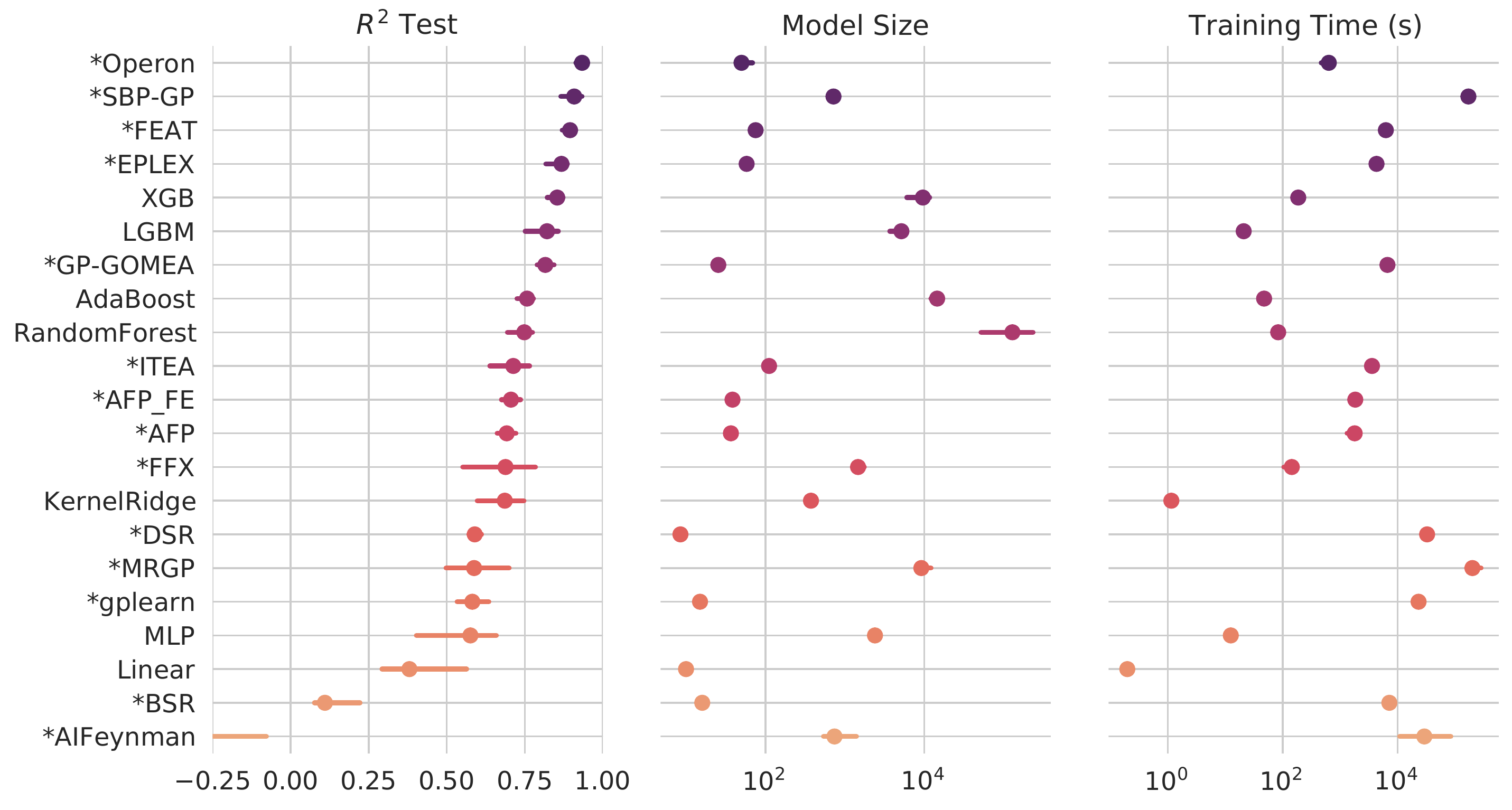}
    \caption{ 
        Results on the black-box regression problems.
        Points indicate the mean of the median test set performance on all problems, and bars show the 95\% confidence interval. 
        Methods marked with an asterisk are SR methods. 
    }
    \label{fig:pmlb_perf}
\end{figure}

\begin{figure}
\begin{minipage}{0.515\textwidth}
        \centering
        \includegraphics[width=\textwidth]{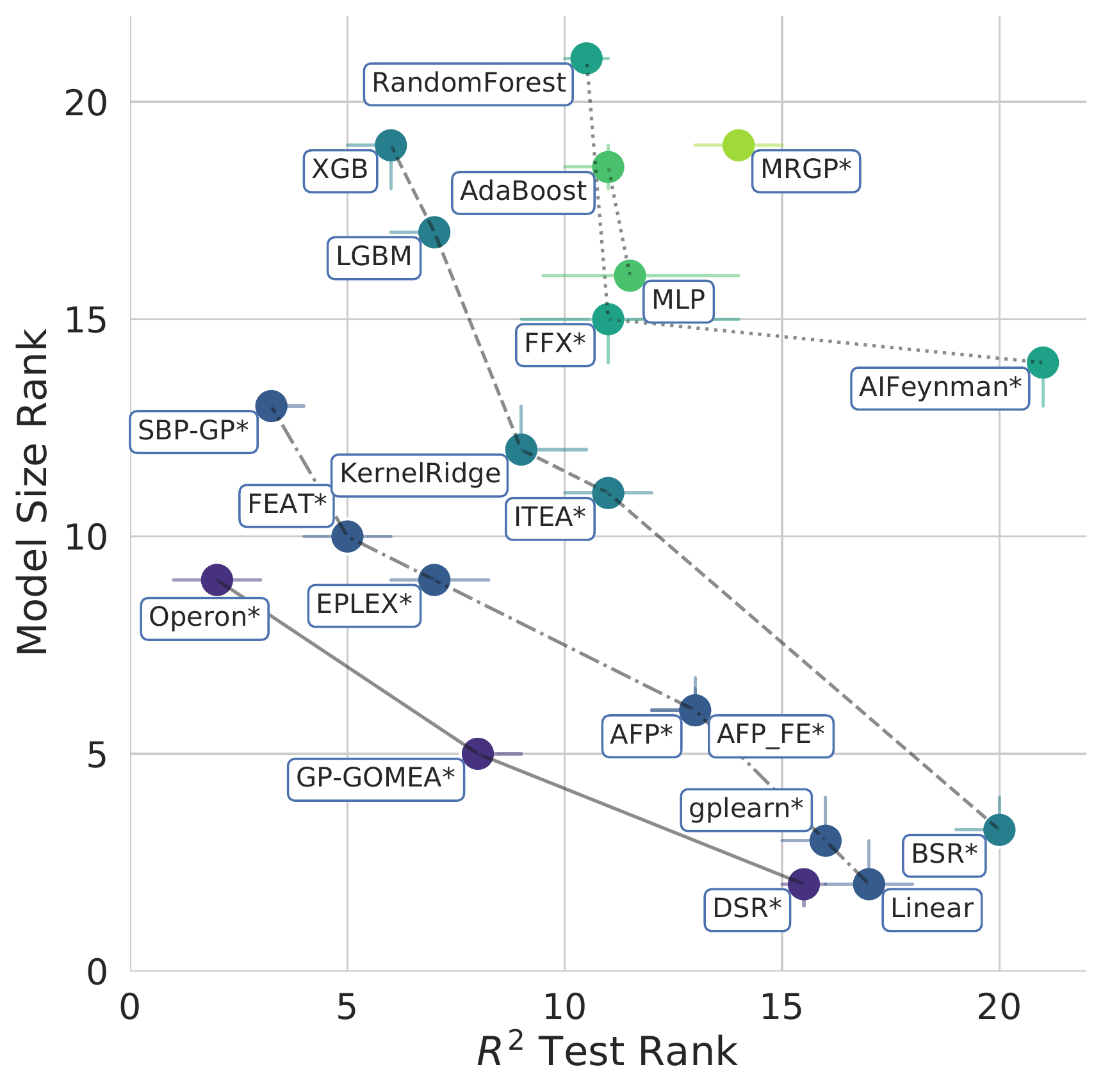}
        \caption{
            Pareto plot comparing the rankings of SR methods in terms of model size and $R^2$ score on the black-box problems.
            Points denote median rankings and the bars denote 95\% confidence intervals. 
            Connecting lines and color denote Pareto dominance rankings. 
            % Pareo sets towards the bottom left are the most efficient trade-offs of model size and accuracy. 
        } \label{fig:pareto}
\end{minipage}
\hspace{0.01\textwidth}
\begin{minipage}{0.475\textwidth}
        \centering
        \includegraphics[width=\textwidth]{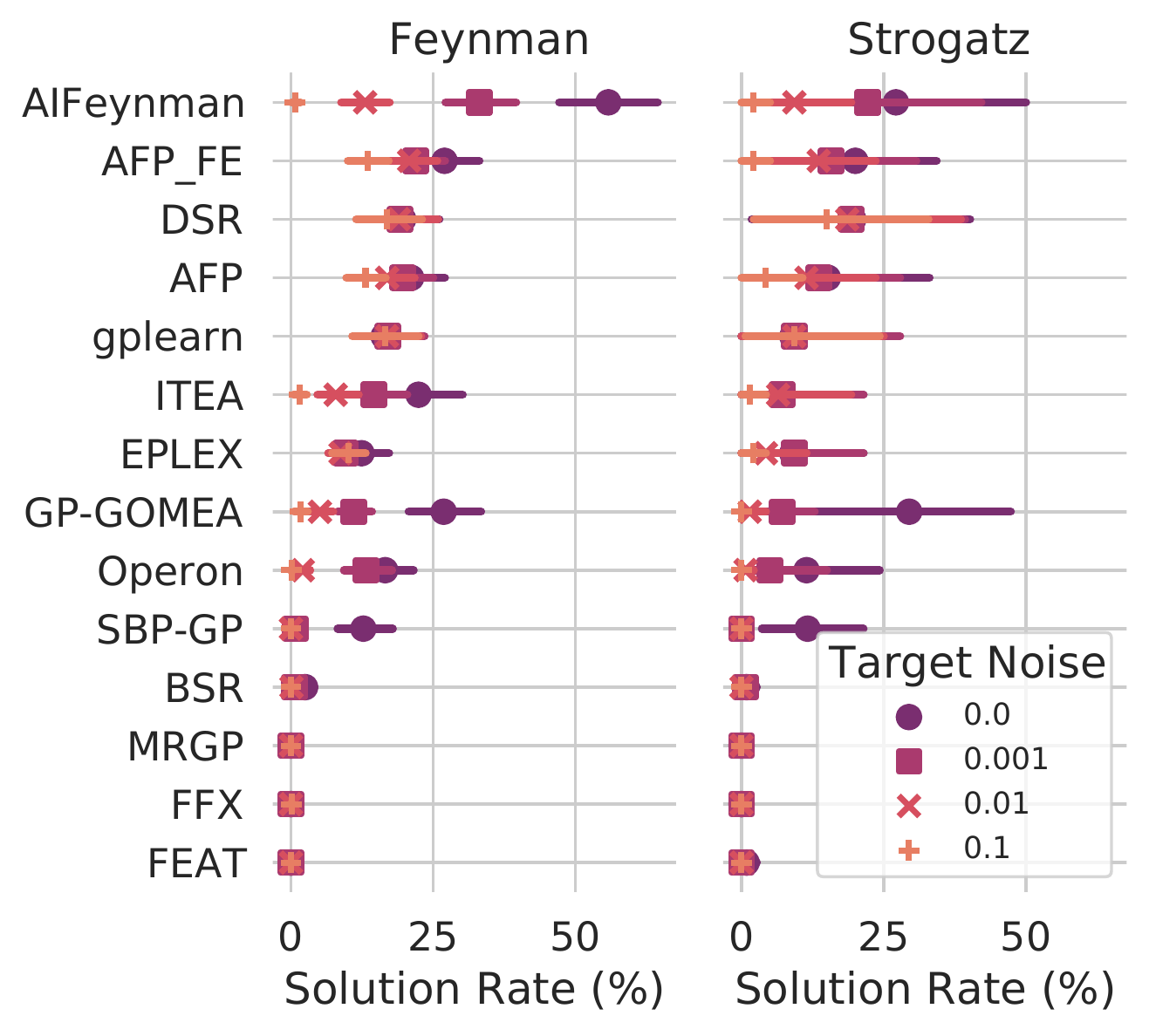}
        \caption{
            Solution rates for the ground-truth regression problems. 
            Color/shape indicates level of noise added to the target variable. 
        }
        \label{fig:symbolic_solns}
\end{minipage}
\end{figure}
%%%%%%%%%%%%%%%%%%%%%%%%%%% RESULTS FIGS %%%%%%%%%%%%%%%%%%%%%%%%%%%%%%%%%%%%%%%

The median test set performance on all problems and methods for the black-box benchmark problems is summarized in Fig.~\ref{fig:pmlb_perf}.  
Across the problems, we find that the models generated by Operon are significantly more accurate than any other method's models in terms of test set $R^2$ ($p\leq${6.5e-05}). 
SBP-GP and FEAT rank second and third and attain similar accuracies, although the models produced by FEAT are significantly smaller ($p=${9.2e-22}).

We note that four of the top five methods (Operon, SBP-GP, FEAT, EPLEX) and six of the top ten methods (GP-GOMEA, ITEA) are GP-based SR methods. 
The other top methods are ensemble tree-based methods, including two popular gradient-boosting algorithms, XGBoost and LightGBM~\cite{chenXgboostScalableTree2016,keLightgbmHighlyEfficient2017}); Random Forest~\cite{breimanRandomForests2001b}; and AdaBoost~\cite{schapireBoostingApproachMachine2003}.
Among these methods, Operon, FEAT and SBP-GP significantly outperform and LightGBM ($p\leq${1.3e-07}) and Operon and SBP-GP outperform XGBoost ($p\leq${1.3e-04}).   
We also note ITEA's overall accuracy is not significantly different from RandomForest or AdaBoost. 
Of note, the models produced by the top five SR methods (aside from SBP-GP) are 1-3 orders of magnitude smaller than models produced by the ensemble tree-based approaches ($p\leq${1.3e-21}).

Among the non-GP-based SR algorithms, FFX and DSR perform similarly to each other ($p=$0.76) and significantly better than BSR and AIFeynman ($p\leq${6.1e-05}).
FFX trains more quickly than DSR, although DSR produces some of the smallest solutions, akin to penalized regression.
We note that AIFeynman performs poorly on these problems, suggesting that not many of them exhibit the qualities of physical systems (rotational/translational invariance, symmetry, etc.) that AIFeynman was designed to exploit. 
Additional statistical comparisons are given in Figs.~\ref{fig:heat_stats_bb}-\ref{fig:heat_stats_soln_sr}.

% pareto curves
In Fig.~\ref{fig:pareto}, we illustrate the performance of the methods on the black-box problems when accuracy and simplicity are considered simultaneously.
The optimal Pareto front for these two objectives (solid line) is composed of three methods: Operon, GP-GOMEA, and DSR, which taken together give the set of best trade-offs between accuracy and simplicity across the black-box regression problems. 
% The Pareto front represents the set of solutions that are pure trade-offs between accuracy and simplicity, in the sense that one cannot identify another method that performs as well or better in one of the two objectives without accepting poorer performance in the other.  

% ground truth regression
Performance on the ground-truth regression problems is summarized in Fig.~\ref{fig:symbolic_solns}, with methods sorted by their median solution rate and grouped by data source (Feynman or Strogatz). 
On average, when the target is free of noise, we observe that AIFeynman identifies exact solutions 53\% of the time, nearly twice as often as the next closest method (GP-GOMEA, 27\%).
However, at noise levels above 0.01, four other methods recover exact solutions more often: DSR, gplearn, AFP\_FE, and AFP.
Taken together, the black-box and ground-truth regression results suggest AIFeynman may be brittle in application to real-world and/or noisy data, yet its performance with little to no noise is significant for the Feynman problems. 
On the Strogatz datasets, AIFeynman's performance is not significantly different than other methods, and indeed there are few significant differences in performance between the top 10 methods at any noise level.
We note that the best method on real-world data, Operon, struggles to recover solutions to these problems, despite finding many candidate solutions with near prefect test set scores. 
See Sec.~\ref{s:add_results}-\ref{s:stats} for additional analysis.
% We see a similar pattern with MRGP, and also note that MRGP exhibits over-fitting to the addition of noise, as noted by the larger final model sizes it produces. 
% \subsection{Black-box Regression Problems}
% overall pmlb results

% \end{wrapfigure}

%%%%%%%%%%%%%%%%%%%%%%%%%%%%%%%%%%%%%%%%%%%%%%%%%%%%%%%%%%%%%%%%%%%%%%%%%%%%%%%%
\section{Discussion and Conclusions}
\label{s:disc}
%%%%%%%%%%%%%%%%%%%%%%%%%%%%%%%%%%%%%%%%%%%%%%%%%%%%%%%%%%%%%%%%%%%%%%%%%%%%%%%%

This paper introduces a SR benchmarking framework that allows objective comparisons of contemporary SR methods on a wide range of diverse regression problems. 
We have found that, on real-world and black-box regression tasks, contemporary GP-based SR methods (e.g. Operon) outperform new SR methods based in other fields of optimization, and can also perform as well as or better than gradient boosted trees while producing simpler models.
On synthetic ground-truth physics and dynamical systems problems, we have verified that AIFeynman finds exact solutions significantly better than other methods when noise is minimal; otherwise, both deep learning-based methods (DSR) and GP-based SR methods (e.g. AFP\_FE) perform best.

% that leverage constant optimization (e.g. Operon, FEAT) 
% We have analyzed 14 contemporary SR methods from multiple angles, including accuracy, solution complexity, running time, ability to discover ground truth solutions, and tolerance of measurement noise.
% By making this benchmark resource open-source, easy to contribute to~\cite{}, and reproducible, we hope to encourage 
%    \item directions for improvement in SR
%        \item running time
%        \item tolerance to noise
%        \item post-run pruning

We see clear ways to improve SRBench by improving the dataset curation, experiment design and analysis. 
For one, we have not benchmarked the methods in a setting that allows them to exploit parallelism, which may change relative run-times. 
There are also many promising SR methods not included in this study that we hope to add in future revisions.  
% % add observations of systems with known dynamics. Caveat: real systems and first principles don't always line up, as in metabolic systems~\cite{yeast} and fluid dynamics \cite{lacavaInferenceCompactNonlinear2016}. 
In addition, whereas our benchmark includes real-world data as well as simulated data with ground-truth models, it does not include real-world data from phenomena with known, first principles models (e.g., observations of a mass-spring-damper system). 
Data such as these could help us better evaluate the ability of SR methods to discover relations under real-world conditions. 
We intend to include these data in future versions, given the evidence that SR models can sometimes discover unexpected analytical models that outperform the expert models in a field (e.g., in studies of yeast metabolism~\cite{schmidtAutomatedRefinementInference2011a} and fluid tank systems~\cite{lacavaInferenceCompactNonlinear2016}). 
% In such cases SR methods may lead to novel insights rather than reproducing known phenomena.
As a final note, our current study highlights orthogonal approaches to SR that show promise, and in future work we hope to explore whether combinations of proposed methods (e.g., non-linear parameter optimization plus semantic search drivers) would have synergistic effects. 
\section*{Acknowledgments}
% Use unnumbered first level headings for the acknowledgments. All acknowledgments
% More information about this disclosure can be found at: \url{https://urldefense.com/v3/__https://neurips.cc/Conferences/2021/PaperInformation/FundingDisclosure__;!!GX6Nv3_Pjr8b-17qtCok029Ok438DqXQ!gLfRVZbnI5EEpjSs2SUEUt3REx61nOcbbaUjq8m9DBea3vF6r9tqUTr1seQz$ }.

% Do {\bf not} include this section in the anonymized submission, only in the final paper. You can use the \texttt{ack} environment provided in the style file to autmoatically hide this section in the anonymized submission.
    William La~Cava was supported by NIH/NLM grant K99-LM012926.
    He would like to thank Curt Calafut and the Penn Medicine Academic Computing Services (PMACS), as well as the PLGrid Infrastructure, for supporting the computational experiments.
    He also thanks members of the Epistasis Lab for their patience, and Joseph D. Romano for coming through in a pinch.   
    
    Ying Jin would like to thank Doctor Jian Guo for hosting an internship for the project and Professor Jian Kang for helpful and inspiring guidance in Bayesian statistics.

    The authors would also like to thank James McDermott for his generous contributions to the repository, and Randal Olson and Weixuan Fu for their initial push to integrate regression benchmarking into PMLB.
    Authors declare no competing interests. 

% \end{ack}

% \section*{References}

% References follow the acknowledgments. Use unnumbered first-level heading for
% the references. Any choice of citation style is acceptable as long as you are
% consistent. It is permissible to reduce the font size to \verb+small+ (9 point)
% when listing the references.
% Note that the Reference section does not count towards the page limit.
\medskip

{
\small

\bibliographystyle{unsrtnat}
\bibliography{SymbolicRegression,additional}
}

%%%%%%%%%%%%%%%%%%%%%%%%%%%%%%%%%%%%%%%%%%%%%%%%%%%%%%%%%%%%
\section*{Checklist}

% %%% BEGIN INSTRUCTIONS %%%
% The checklist follows the references.  Please
% read the checklist guidelines carefully for information on how to answer these
% questions.  For each question, change the default \answerTODO{} to \answerYes{},
% \answerNo{}, or \answerNA{}.  You are strongly encouraged to include a {\bf
% justification to your answer}, either by referencing the appropriate section of
% your paper or providing a brief inline description.  For example:
% \begin{itemize}
%   \item Did you include the license to the code and datasets? \answerYes{See Section~\ref{gen_inst}.}
%   \item Did you include the license to the code and datasets? \answerNo{The code and the data are proprietary.}
%   \item Did you include the license to the code and datasets? \answerNA{}
% \end{itemize}
% Please do not modify the questions and only use the provided macros for your
% answers.  Note that the Checklist section does not count towards the page
% limit.  In your paper, please delete this instructions block and only keep the
% Checklist section heading above along with the questions/answers below.
%%% END INSTRUCTIONS %%%

\begin{enumerate}

\item For all authors...
\begin{enumerate}
  \item Do the main claims made in the abstract and introduction accurately reflect the paper's contributions and scope?
    \answerYes{The results can be verified by visiting our repository. 
               Specific claims are supported by statistical tests.}
  \item Did you describe the limitations of your work?
    \answerYes{See discussion and conclusions.}
  \item Did you discuss any potential negative societal impacts of your work?
    \answerYes{See Sec.~\ref{s:ethics}.}
  \item Have you read the ethics review guidelines and ensured that your paper conforms to them?
    \answerYes{In addition to releasing the benchmark in a transparent way, we discuss ethics in Sec.~\ref{s:ethics}.}
\end{enumerate}

% \item If you are including theoretical results...
% \begin{enumerate}
%   \item Did you state the full set of assumptions of all theoretical results?
%     \answerTODO{}
% 	\item Did you include complete proofs of all theoretical results?
%     \answerTODO{}
% \end{enumerate}

\item If you ran experiments (e.g. for benchmarks)...
\begin{enumerate}
  \item Did you include the code, data, and instructions needed to reproduce the main experimental results (either in the supplemental material or as a URL)?
    \answerYes{See \url{https://github.com/EpistasisLab/srbench}.}
  \item Did you specify all the training details (e.g., data splits, hyperparameters, how they were chosen)?
    \answerYes{See Table~\ref{tbl:exp}. }
	\item Did you report error bars (e.g., with respect to the random seed after running experiments multiple times)?
    \answerYes{See Figs.~\ref{fig:pmlb_perf}-\ref{fig:symbolic_solns} for example.}
	\item Did you include the total amount of compute and the type of resources used (e.g., type of GPUs, internal cluster, or cloud provider)?
    \answerYes{see Appendix}
\end{enumerate}

\item If you are using existing assets (e.g., code, data, models) or curating/releasing new assets...
\begin{enumerate}
  \item If your work uses existing assets, did you cite the creators?
    \answerYes{}
  \item Did you mention the license of the assets?
    \answerYes{}
  \item Did you include any new assets either in the supplemental material or as a URL?
    \answerYes{}
  \item Did you discuss whether and how consent was obtained from people whose data you're using/curating?
    \answerYes{Datasets are released under an MIT license.}
  \item Did you discuss whether the data you are using/curating contains personally identifiable information or offensive content?
    \answerYes{These datasets do not contain personally identifiable information.}
\end{enumerate}

\end{enumerate}

%%%%%%%%%%%%%%%%%%%%%%%%%%%%%%%%%%%%%%%%%%%%%%%%%%%%%%%%%%%%
\newpage

\appendix

%TODOS
% PR for feynman PMLB
% running time fig
% cross pollination background from kommenda

\section{Appendix}
\label{s:appendix}

Please refer to \url{https://github.com/EpistasisLab/srbench/} for the most up-to-date guide to SRBench. 

\subsection{Running the Benchmark}
\label{s:howto}

The README in our Github repository includes the set of commands to reproduce the benchmark experiment, which are summarized here. 
Experiments are launched from the \texttt{experiments/} folder via the script \texttt{analyze.py}.
The script can be configured to run the experiment in parallel locally, on an LSF job scheduler, or on a SLURM job scheduler. 
To see the full set of options, run \texttt{python analyze.py -h}. 

After installing and configuring the conda environment, the complete black-box experiment can be started via the command:
\\
\begin{quote}
\texttt{
python analyze.py /path/to/pmlb/datasets -n\_trials 10 -results ../results -time\_limit 48:00
}
\end{quote}

Similarly, the ground-truth regression experiment for Strogatz datasets and a target noise of 0.0 are run by the command:
\\
\begin{quote}
\texttt{
python analyze.py -results ../results\_sym\_data -target\_noise 0.0 "/path/to/pmlb/datasets/strogatz*" -sym\_data -n\_trials 10 -time\_limit 9:00 -tuned
}
\end{quote}

\subsection{Contributing a Method}
A living version of the method contribution instructions are described in the \href{https://github.com/EpistasisLab/srbench/blob/master/CONTRIBUTING.md}{Contribution Guide}.
To illustrate the simplicity of contributing a method, Figure~\ref{fig:ex_code} shows the script submitted for Bayesian Symbolic Regression~\cite{jinBayesianSymbolicRegression2020}. 
In addition to the code snippet, authors may either add their code package to the conda/pip environment, or provide an install script.
When a pull request is issued by a contributor, new methods and installs are automatically tested on a minimal version of the benchmark.  
Once the tests pass and the method is approved by the benchmark maintainers, the contribution becomes part of the resource and can be tested via the commands above. 

\begin{figure}
	\input{contribution_example}
    \caption{
        An example code contribution, defining the estimator, its hyperparameters, and functions to return the complexity and symbolic model.
    }\label{fig:ex_code} 
\end{figure}

\subsection{Additional Background and Motivation}
\label{s:add_background}

% cross pollination
% DSR showed good performance in finding exact solutions to a number of synthetic benchmark problems, although 6 of the 12 problems reported by the authors  (c.f. Table 1,~\cite{petersenDeepSymbolicRegression2020}) are low-order synthetic polynomials that were proposed for black-listing in community surveys (c.f. Table 3, \citet{whiteBetterGPBenchmarks2012a}).  

\paragraph{Eureqa}
Eureqa is a commercial GP-based SR software that was acquired by DataRobot in 2017\footnote{\url{https://www.datarobot.com/nutonian/}}. 
Due to its closed-source nature and incorporation into the DataRobot platform, it is impossible to benchmark its performance while controlling for important experimental variables such as number of evaluations, space and time limits, population size, and so forth. 
However, the novel algorithmic aspects of Eureqa are rooted in a number of ablation studies~\cite{schmidtComparisonTreeGraph2007,schmidtCoevolutionFitnessPredictors2008,schmidtAgefitnessParetoOptimization2011} that we summarize here. 
First is its use of directed acyclic graphs for representing equations in lieu of trees, which resulted in more space-efficient model encoding relative to trees, without a significant difference in accuracy~\cite{schmidtComparisonTreeGraph2007}. 
The most significant improvement over traditional tournament-based selection is Eureqa's use of age-fitness Pareto optimization (AFP), a method in which random restarts are incorporated each generation as new offspring, and are protected from competing with older, more fit equations by including age as an objective to be minimized~\cite{schmidtAgefitnessParetoOptimization2011}. 
Eureqa also includes the co-evolution of fitness predictors, in which fitness assignment is sped up by optimizing a second population of training sample indices that best distinguish between equations in the population~\cite{schmidtCoevolutionFitnessPredictors2008}.
Unfortunately we cannot guarantee that Eureqa currently uses any of these reported algorithms for SR, due to its closed-source nature.
We chose instead to benchmark known algorithms (AFP, AFP\_FE) with open-source implementations, hoping that the resulting study's conclusions may better inform future methods development.
We note that AFP has been outperformed by a number of other optimization methods in controlled studies since its release (e.g.,~\cite{lacavaEpsilonLexicaseSelectionRegression2016c,liskowskiDiscoverySearchObjectives2017}). 

\paragraph{Constant optimization in Genetic Programming}
One of the clearest improvements over Koza-style GP has been the adoption of local search methods to handle constant optimization distinctly from evolutionary learning. 
Regarding the optimization of constants in GP, several reasons can explain why backpropagation and gradient descent can be considered to be relatively under-used in GP (compared to, e.g., evolutionary neural architecture search). For example, early works often ignored the use of feature standardization (e.g., by z-scoring), the lack of which can harm gradient propagation~\cite{dickFeatureStandardisationCoefficient2020}. Next to this, GP relies on crafting compositions out of a multitude of operations, some of which are prone to cause vanishing or exploding gradients. Last but not least, to the best of our knowledge, the field lacks a comprehensive study that provides guidelines for the appropriate hyperparameters for constant optimization (learning rate schedule, iterations, batch size, etc.), and how to effectively balance parameter learning with the evolutionary process.

% \subsection{Extended Methods}

\subsection{Additional Dataset Information}
\label{s:add_dataset}

All datasets, including metadata, are available from \href{https://epistasislab.github.io/pmlb/}{PMLB}. 
Each dataset is stored using Git Large File Storage and PMLB is planned for long-term maintenance.
PMLB is available under an MIT license, and is described in detail in~\citet{romanoPMLBV1Open2021}. 
The authors bear all responsibility in case of violation of rights.

\begin{figure}
    \centering
    \includegraphics[width=0.7\textwidth]{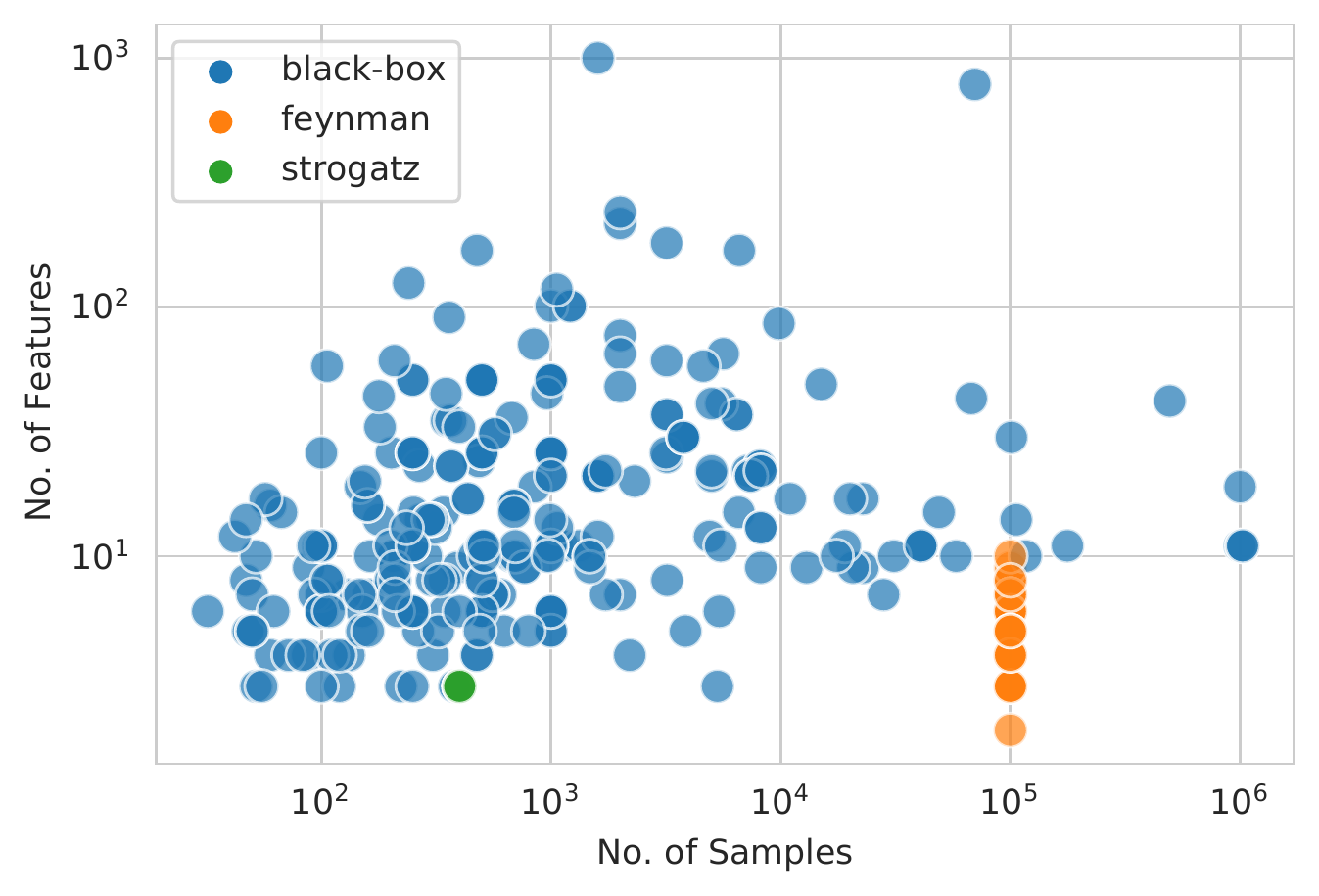}
    \caption{Distribution of dataset sizes in PMLB.}
    \label{fig:pmlb}
\end{figure}

\paragraph{Dataset Properties}
The distribution of dataset sizes by samples and features are shown in Fig.~\ref{fig:pmlb}. 
Datasets vary in size from tens to millions of samples, and up to thousands of features. 
The datasets can be navigated and inspected in the \href{https://epistasislab.github.io/pmlb/}{repository documentation}.

\paragraph{Ethical Considerations and Intended Uses}\label{s:ethics}
PMLB is intended to be used as a framework for benchmarking ML and SR algorithms and as a resource for investigating the structure of datasets. 
This paper does not contribute new datasets, but rather collates and standardizes datasets that were already publicly available.
In that regard, we do not foresee SRBench as creating additional ethical issues around their use.
Nevertheless, it is worth noting that PMLB contains well-known, real-world datasets from UCI and OpenML for which ethical considerations are important, such as the \href{https://github.com/EpistasisLab/pmlb/blob/master/datasets/1089_USCrime/metadata.yaml}{USCrime} dataset. 
Whereas we would view the risk of harm arising specifically from this dataset to be low (the data is from 1960), it is exemplary of a task for which algorithmic decision making could exacerbate existing biases in the criminal justice system.  
As such it is used as a benchmark in a number of papers in the ML fairness literature (e.g.~\cite{kearnsPreventingFairnessGerrymandering2018, lacavaGeneticProgrammingApproaches2020}). 
None of the datasets herein contain personally identifiable information.

\paragraph{Feynman datasets}
The Feynman benchmarks were sourced from the \href{https://space.mit.edu/home/tegmark/aifeynman.html}{Feynman Symbolic Regression Database}. 
We standardized the Feynman and Bonus equations to PMLB format and included metadata detailing the model form and the units for each variable. 
We used the version of the equations that were not simplified by dimensional analysis. 
\citet{udrescuAIFeynmanPhysicsInspired2020} describe each dataset as containing $10^5$ rows, but each actually contains $10^6$. 
Given this discrepancy and after noting that sub-sampling did not significantly change the correlation structure of any of the problems, each dataset was down-sampled from 1 million samples to 100,000 to lower the computational burden.
We also observed that Eqn. II.11.17 was missing from the database. 
Finally, we excluded three datasets from our analysis that contained $\arcsin$ and $\arccos$ functions, as these were not implemented in the majority of SR algorithms we tested.

\paragraph{Strogatz datasets}
The Strogatz datasets were sourced from the \href{https://github.com/lacava/ode-strogatz}{ODE-Strogatz repository}~\cite{lacavaInferenceCompactNonlinear2016}.
Each dataset is one state of a 2-state system of first-order, ordinary differential equations (ODEs). 
The goal of each problem is to predict rate of change of the state given the current two states on which it depends. 
Each represents natural processes that exhibit chaos and non-linear dynamics.
The problems were originally adapted from~\cite{strogatzNonlinearDynamicsChaos2014} by~\citet{schmidtMachineScienceAutomated2011}.   
In order to simulate their behavior, initial conditions were chosen within stable basins of attraction.
Each system was simulated using Simulink, and the simulation code is available in the repository above.
The equations for each of these datasets are shown in Table~\ref{tbl:strogatz}.

\begin{table}[htb]
\scriptsize
\centering
\renewcommand{\arraystretch}{1.2}\addtolength{\tabcolsep}{.5pt}
\caption{The Strogatz ODE problems. 
         } \label{tbl:strogatz}
% \rowcolors{3}{white}{lightgray}
\begin{tabular}{ll} %L{0.12\textwidth} L{0.25\textwidth} L{0.25\textwidth} L{0.25\textwidth} } \toprule 
 Name & Target \\ \toprule 
Bacterial Respiration                   & $\dot{x} = 20 - x - \frac{x \cdot y}{1+0.5 \cdot x^2}$ \\	
                                        & $ \dot{y} = 10 - \frac{x \cdot y}{1+0.5 \cdot x^2}$	\\ 
\midrule
Bar Magnets                             & $\dot{\theta} = 0.5 \cdot \sin (\theta - \phi) - \sin (\theta)$	\\
                                        & $ \dot{\phi} = 0.5 \cdot \sin (\phi - \theta) - \sin (\phi)$ \\ 
\midrule
Glider                                  & $\dot{v} = - 0.05 \cdot  v^2 - sin (\theta)$	\\
                                        & $ \dot{\theta} = v - \cos (\theta)/v$	\\ 
\midrule
Lotka-Volterra interspecies dynamics    & $\dot{x} = 3  \cdot x - 2  \cdot x \cdot y - x^2$	\\
                                        & $ \dot{y} = 2 \cdot y - x \cdot y - y^2$	\\ 
\midrule
Predator Prey                           & $\dot{x} = x  \cdot \left( 4 - x - \frac{y}{1+x} \right)$	\\
                                        & $\dot{y} = y \cdot \left( \frac{x}{1+x} - 0.075 \cdot y \right)$	\\ 
\midrule
Shear Flow                              & $\dot{\theta} = \cot (\phi) \cdot cos(\theta)$	\\
                                        & $ \dot{\phi} = \left(\cos ^2 (\phi) + 0.1 \cdot  \sin^2 (\phi)\right) \cdot sin(\theta)$	\\ 
\midrule
van der Pol oscillator                  & $\dot{x} = 10 \cdot  \left(y - \frac{1}{3} \cdot (x^3-x) \right)$	\\
                                        & $ \dot{y} = -\frac{1}{10} \cdot x$ \\
\bottomrule
\end{tabular}
\end{table}

\paragraph{Adding Noise}
White gaussian noise was added to the target as a fraction of the signal root mean square value. 
In other words, for target noise level $\gamma$,

\[
    y_{noise} = y + \epsilon,\; \epsilon \sim \mathcal{N} \left( 0, \gamma \sqrt{ \frac{1}{N}\sum{y_i^2} } \right)
\]

\subsection{Additional Experiment Details}
\label{s:add_exp}

%% LPC
Experiments were run in a heterogeneous cluster computing environment composed of hosts with 24-28 core Intel(R) Xeon(R) CPU E5-2690 v4 @ 2.60GHz processors and 250 GB of RAM. 
Jobs consisted of the training of each method on a single dataset for a fixed random seed. 
Each job received one CPU core and up to 16GB of RAM, and was time-limited as shown in Table~\ref{tbl:exp}. 
For the ground-truth problems, the final models from each method were given an additional hour of computing time with 8GB of RAM to be simplified with sympy and assessed by the solution criteria (see Def.~\ref{def:soln}).
For the black-box problems, if a job was killed due to the time limit, we re-ran the experiment without hyperparameter tuning, thereby only requiring a single training iteration to complete within 48 hours. 
To ease the computational burden for large datasets, training data exceeding 10,000 samples was randomly subset to 10,000 rows; test set predictions were still evaluated over the entire test fold.  

The hyperparameter settings for each method are shown in Tables~\ref{tbl:ml_methods}-\ref{tbl:sr_methods2}. 
Each SR method was tuned from a set of six hyperparameter combinations. 
The most common parameter setting chosen during the black-box regression experiments was then used as the ``tuned" version of each algorithm for the ground-truth problems, with updates to 1) include any mathematical operators needed for those problems and 2) double the evaluation budget. 

%% Hyperparameters of each method
\begin{table}
    \footnotesize
    \centering

    \caption{
        ML methods and the hyperparameter spaces used in tuning.
    }
	\label{tbl:ml_methods}
    \rowcolors{2}{gray!10}{white}
    \begin{tabular}{l p{37em}}
\toprule
          Method &                                                                                                                                                          Hyperparameters \\
\midrule
        AdaBoost &                                                                                               \{'learning\_rate': (0.01, 0.1, 1.0, 10.0), 'n\_estimators': (10, 100, 1000)\} \\
     KernelRidge &                                                           \{'kernel': ('linear', 'poly', 'rbf', 'sigmoid'), 'alpha': (0.0001, 0.01, 0.1, 1), 'gamma': (0.01, 0.1, 1, 10)\} \\
       LassoLars &                                                                                                                                 \{'alpha': (0.0001, 0.001, 0.01, 0.1, 1)\} \\
            LGBM & \{'n\_estimators': (10, 50, 100, 250, 500, 1000), 'learning\_rate': (0.0001, 0.01, 0.05, 0.1, 0.2), 'subsample': (0.5, 0.75, 1), 'boosting\_type': ('gbdt', 'dart', 'goss')\} \\
LinearRegression &                                                                                                                                               \{'fit\_intercept': (True,)\} \\
             MLP &                                \{'activation': ('logistic', 'tanh', 'relu'), 'solver': ('lbfgs', 'adam', 'sgd'), 'learning\_rate': ('constant', 'invscaling', 'adaptive')\} \\
    RandomForest &                                                  \{'n\_estimators': (10, 100, 1000), 'min\_weight\_fraction\_leaf': (0.0, 0.25, 0.5), 'max\_features': ('sqrt', 'log2', None)\} \\
             SGD &                                                                                               \{'alpha': (1e-06, 0.0001, 0.01, 1), 'penalty': ('l2', 'l1', 'elasticnet')\} \\
             XGB &          \{'n\_estimators': (10, 50, 100, 250, 500, 1000), 'learning\_rate': (0.0001, 0.01, 0.05, 0.1, 0.2), 'gamma': (0, 0.1, 0.2, 0.3, 0.4), 'subsample': (0.5, 0.75, 1)\} \\
\bottomrule
\end{tabular}

\end{table}

\begin{table}
    \scriptsize
    \centering
    \caption{
        Part 1: SR methods and the hyperparameter spaces used in tuning on the black-box regression problems.
    }
	\label{tbl:sr_methods1}
    \begin{tabular}{l p{37em}}
\toprule
   Method &                                                                                                      Hyperparameters \\
\midrule
\midrule
      AFP &               \{'popsize': 100, 'g': 2500, 'op\_list': ['n', 'v', '+', '-', '*', '/', 'exp', 'log', '2', '3', 'sqrt']\} \\
          & \{'popsize': 100, 'g': 2500, 'op\_list': ['n', 'v', '+', '-', '*', '/', 'exp', 'log', '2', '3', 'sqrt', 'sin', 'cos']\} \\
          &                \{'popsize': 500, 'g': 500, 'op\_list': ['n', 'v', '+', '-', '*', '/', 'exp', 'log', '2', '3', 'sqrt']\} \\
          &  \{'popsize': 500, 'g': 500, 'op\_list': ['n', 'v', '+', '-', '*', '/', 'exp', 'log', '2', '3', 'sqrt', 'sin', 'cos']\} \\
          &               \{'popsize': 1000, 'g': 250, 'op\_list': ['n', 'v', '+', '-', '*', '/', 'exp', 'log', '2', '3', 'sqrt']\} \\
          & \{'popsize': 1000, 'g': 250, 'op\_list': ['n', 'v', '+', '-', '*', '/', 'exp', 'log', '2', '3', 'sqrt', 'sin', 'cos']\} \\
\midrule
AIFeynman &    \{'BF\_try\_time': 60, 'NN\_epochs': 4000, 'BF\_ops\_file\_type'="10ops.txt"\} \\
          &    \{'BF\_try\_time': 60, 'NN\_epochs': 4000, 'BF\_ops\_file\_type'="14ops.txt"\} \\
          &    \{'BF\_try\_time': 60, 'NN\_epochs': 4000, 'BF\_ops\_file\_type'="19ops.txt"\} \\
          &    \{'BF\_try\_time': 600, 'NN\_epochs': 400, 'BF\_ops\_file\_type'="10ops.txt"\} \\
          &    \{'BF\_try\_time': 600, 'NN\_epochs': 400, 'BF\_ops\_file\_type'="14ops.txt"\} \\
          &    \{'BF\_try\_time': 600, 'NN\_epochs': 400, 'BF\_ops\_file\_type'="19ops.txt"\} \\
\midrule
      BSR &                                                                           \{'treeNum': 6, 'itrNum': 500, 'val': 1000\} \\
          &                                                                           \{'treeNum': 6, 'itrNum': 1000, 'val': 500\} \\
          &                                                                           \{'treeNum': 3, 'itrNum': 500, 'val': 1000\} \\
          &                                                                           \{'treeNum': 6, 'itrNum': 5000, 'val': 100\} \\
          &                                                                           \{'treeNum': 3, 'itrNum': 5000, 'val': 100\} \\
          &                                                                           \{'treeNum': 3, 'itrNum': 1000, 'val': 500\} \\
\midrule
      DSR &                                                      \{'batch\_size': array([    10,    100,   1000,  10000, 100000])\} \\
\midrule
    EPLEX &               \{'popsize': 1000, 'g': 250, 'op\_list': ['n', 'v', '+', '-', '*', '/', 'exp', 'log', '2', '3', 'sqrt']\} \\
          &                \{'popsize': 500, 'g': 500, 'op\_list': ['n', 'v', '+', '-', '*', '/', 'exp', 'log', '2', '3', 'sqrt']\} \\
          & \{'popsize': 1000, 'g': 250, 'op\_list': ['n', 'v', '+', '-', '*', '/', 'sin', 'cos', 'exp', 'log', '2', '3', 'sqrt']\} \\
          &               \{'popsize': 100, 'g': 2500, 'op\_list': ['n', 'v', '+', '-', '*', '/', 'exp', 'log', '2', '3', 'sqrt']\} \\
          & \{'popsize': 100, 'g': 2500, 'op\_list': ['n', 'v', '+', '-', '*', '/', 'sin', 'cos', 'exp', 'log', '2', '3', 'sqrt']\} \\
          &  \{'popsize': 500, 'g': 500, 'op\_list': ['n', 'v', '+', '-', '*', '/', 'sin', 'cos', 'exp', 'log', '2', '3', 'sqrt']\} \\
\midrule
     FEAT &                                                                           \{'pop\_size': 100, 'gens': 2500, 'lr': 0.1\} \\
          &                                                                           \{'pop\_size': 100, 'gens': 2500, 'lr': 0.3\} \\
          &                                                                            \{'pop\_size': 500, 'gens': 500, 'lr': 0.1\} \\
          &                                                                            \{'pop\_size': 500, 'gens': 500, 'lr': 0.3\} \\
          &                                                                           \{'pop\_size': 1000, 'gens': 250, 'lr': 0.1\} \\
          &                                                                           \{'pop\_size': 1000, 'gens': 250, 'lr': 0.3\} \\
\midrule
   FE\_AFP & \{'popsize': 1000, 'g': 250, 'op\_list': ['n', 'v', '+', '-', '*', '/', 'sin', 'cos', 'exp', 'log', '2', '3', 'sqrt']\} \\
          &                \{'popsize': 500, 'g': 500, 'op\_list': ['n', 'v', '+', '-', '*', '/', 'exp', 'log', '2', '3', 'sqrt']\} \\
          &               \{'popsize': 1000, 'g': 250, 'op\_list': ['n', 'v', '+', '-', '*', '/', 'exp', 'log', '2', '3', 'sqrt']\} \\
          &               \{'popsize': 100, 'g': 2500, 'op\_list': ['n', 'v', '+', '-', '*', '/', 'exp', 'log', '2', '3', 'sqrt']\} \\
          & \{'popsize': 100, 'g': 2500, 'op\_list': ['n', 'v', '+', '-', '*', '/', 'sin', 'cos', 'exp', 'log', '2', '3', 'sqrt']\} \\
          &  \{'popsize': 500, 'g': 500, 'op\_list': ['n', 'v', '+', '-', '*', '/', 'sin', 'cos', 'exp', 'log', '2', '3', 'sqrt']\} \\
\bottomrule
\end{tabular}

\end{table}

\begin{table}
    \scriptsize

    \centering

    \caption{
        Part 2: SR methods and the hyperparameter spaces used in tuning on the black-box regression problems.
    }
	\label{tbl:sr_methods2}
    \begin{tabular}{l p{37em}}
\toprule
       Method &                                                                                                                                                                                                                                                                                Hyperparameters \\
\midrule
\midrule
      GPGOMEA &                                                                                                                                                                        \{'initmaxtreeheight': (4,), 'functions': ('+\_-\_*\_p/\_plog\_sqrt\_sin\_cos',), 'popsize': (1000,), 'linearscaling': (True,)\} \\
              &                                                                                                                                                                        \{'initmaxtreeheight': (6,), 'functions': ('+\_-\_*\_p/\_plog\_sqrt\_sin\_cos',), 'popsize': (1000,), 'linearscaling': (True,)\} \\
              &                                                                                                                                                                                          \{'initmaxtreeheight': (4,), 'functions': ('+\_-\_*\_p/',), 'popsize': (1000,), 'linearscaling': (True,)\} \\
              &                                                                                                                                                                                          \{'initmaxtreeheight': (6,), 'functions': ('+\_-\_*\_p/',), 'popsize': (1000,), 'linearscaling': (True,)\} \\
              &                                                                                                                                                                       \{'initmaxtreeheight': (4,), 'functions': ('+\_-\_*\_p/\_plog\_sqrt\_sin\_cos',), 'popsize': (1000,), 'linearscaling': (False,)\} \\
              &                                                                                                                                                                       \{'initmaxtreeheight': (6,), 'functions': ('+\_-\_*\_p/\_plog\_sqrt\_sin\_cos',), 'popsize': (1000,), 'linearscaling': (False,)\} \\
\midrule
         ITEA &                                                                                                                                                                             \{'exponents': ((-5, 5),), 'termlimit': ((2, 15),), 'transfunctions': ('[Id, Tanh, Sin, Cos, Log, Exp, SqrtAbs]',)\} \\
              &                                                                                                                                                                              \{'exponents': ((-5, 5),), 'termlimit': ((2, 5),), 'transfunctions': ('[Id, Tanh, Sin, Cos, Log, Exp, SqrtAbs]',)\} \\
              &                                                                                                                                                                                                           \{'exponents': ((-5, 5),), 'termlimit': ((2, 15),), 'transfunctions': ('[Id, Sin]',)\} \\
              &                                                                                                                                                                                                            \{'exponents': ((0, 5),), 'termlimit': ((2, 15),), 'transfunctions': ('[Id, Sin]',)\} \\
              &                                                                                                                                                                                                             \{'exponents': ((0, 5),), 'termlimit': ((2, 5),), 'transfunctions': ('[Id, Sin]',)\} \\
              &                                                                                                                                                                              \{'exponents': ((0, 5),), 'termlimit': ((2, 15),), 'transfunctions': ('[Id, Tanh, Sin, Cos, Log, Exp, SqrtAbs]',)\} \\
\midrule
         MRGP &                                                                                                                                                                                                                                    \{'popsize': 1000, 'g': 250, 'rt\_cross': 0.8, 'rt\_mut': 0.2\} \\
              &                                                                                                                                                                                                                                    \{'popsize': 100, 'g': 2500, 'rt\_cross': 0.2, 'rt\_mut': 0.8\} \\
              &                                                                                                                                                                                                                                    \{'popsize': 100, 'g': 2500, 'rt\_cross': 0.8, 'rt\_mut': 0.2\} \\
              &                                                                                                                                                                                                                                     \{'popsize': 500, 'g': 500, 'rt\_cross': 0.2, 'rt\_mut': 0.8\} \\
              &                                                                                                                                                                                                                                     \{'popsize': 500, 'g': 500, 'rt\_cross': 0.8, 'rt\_mut': 0.2\} \\
              &                                                                                                                                                                                                                                    \{'popsize': 1000, 'g': 250, 'rt\_cross': 0.2, 'rt\_mut': 0.8\} \\
\midrule
       Operon &                  \{'population\_size': (500,), 'pool\_size': (500,), 'max\_length': (50,), 'allowed\_symbols': ('add,mul,aq,constant,variable',), 'local\_iterations': (5,), 'offspring\_generator': ('basic',), 'tournament\_size': (5,), 'reinserter': ('keep-best',), 'max\_evaluations': (500000,)\} \\
              & \{'population\_size': (500,), 'pool\_size': (500,), 'max\_length': (25,), 'allowed\_symbols': ('add,mul,aq,exp,log,sin,tanh,constant,variable',), 'local\_iterations': (5,), 'offspring\_generator': ('basic',), 'tournament\_size': (5,), 'reinserter': ('keep-best',), 'max\_evaluations': (500000,)\} \\
              &                  \{'population\_size': (500,), 'pool\_size': (500,), 'max\_length': (25,), 'allowed\_symbols': ('add,mul,aq,constant,variable',), 'local\_iterations': (5,), 'offspring\_generator': ('basic',), 'tournament\_size': (5,), 'reinserter': ('keep-best',), 'max\_evaluations': (500000,)\} \\
              &                  \{'population\_size': (100,), 'pool\_size': (100,), 'max\_length': (50,), 'allowed\_symbols': ('add,mul,aq,constant,variable',), 'local\_iterations': (5,), 'offspring\_generator': ('basic',), 'tournament\_size': (3,), 'reinserter': ('keep-best',), 'max\_evaluations': (500000,)\} \\
              & \{'population\_size': (100,), 'pool\_size': (100,), 'max\_length': (25,), 'allowed\_symbols': ('add,mul,aq,exp,log,sin,tanh,constant,variable',), 'local\_iterations': (5,), 'offspring\_generator': ('basic',), 'tournament\_size': (3,), 'reinserter': ('keep-best',), 'max\_evaluations': (500000,)\} \\
              &                  \{'population\_size': (100,), 'pool\_size': (100,), 'max\_length': (25,), 'allowed\_symbols': ('add,mul,aq,constant,variable',), 'local\_iterations': (5,), 'offspring\_generator': ('basic',), 'tournament\_size': (3,), 'reinserter': ('keep-best',), 'max\_evaluations': (500000,)\} \\
\midrule
      gplearn &                                                                                                                                                                                     \{'population\_size': 100, 'generations': 5000, 'function\_set': ('add', 'sub', 'mul', 'div', 'log', 'sqrt')\} \\
              &                                                                                                                                                                                     \{'population\_size': 1000, 'generations': 500, 'function\_set': ('add', 'sub', 'mul', 'div', 'log', 'sqrt')\} \\
              &                                                                                                                                                                       \{'population\_size': 1000, 'generations': 500, 'function\_set': ('add', 'sub', 'mul', 'div', 'log', 'sqrt', 'sin', 'cos')\} \\
              &                                                                                                                                                                                     \{'population\_size': 500, 'generations': 1000, 'function\_set': ('add', 'sub', 'mul', 'div', 'log', 'sqrt')\} \\
              &                                                                                                                                                                       \{'population\_size': 500, 'generations': 1000, 'function\_set': ('add', 'sub', 'mul', 'div', 'log', 'sqrt', 'sin', 'cos')\} \\
              &                                                                                                                                                                       \{'population\_size': 100, 'generations': 5000, 'function\_set': ('add', 'sub', 'mul', 'div', 'log', 'sqrt', 'sin', 'cos')\} \\
\midrule
sembackpropgp &                                                                                                 \{'popsize': (1000,), 'functions': ('+\_-\_*\_aq\_plog\_sin\_cos',), 'linearscaling': (False,), 'sbrdo': (0.9,), 'submut': (0.1,), 'tournament': (4,), 'maxsize': (250,), 'sblibtype': ('p\_6\_9999',)\} \\
              &                                                                                                                             \{'popsize': (1000,), 'functions': ('+\_-\_*\_aq\_plog\_sin\_cos',), 'linearscaling': (True,), 'sbrdo': (0.9,), 'submut': (0.1,), 'tournament': (4,), 'maxsize': (1000,)\} \\
              &                                                                                                                             \{'popsize': (1000,), 'functions': ('+\_-\_*\_aq\_plog\_sin\_cos',), 'linearscaling': (True,), 'sbrdo': (0.9,), 'submut': (0.1,), 'tournament': (8,), 'maxsize': (1000,)\} \\
              &                                                                                                                             \{'popsize': (1000,), 'functions': ('+\_-\_*\_aq\_plog\_sin\_cos',), 'linearscaling': (True,), 'sbrdo': (0.9,), 'submut': (0.1,), 'tournament': (4,), 'maxsize': (5000,)\} \\
              &                                                                                                                             \{'popsize': (1000,), 'functions': ('+\_-\_*\_aq\_plog\_sin\_cos',), 'linearscaling': (True,), 'sbrdo': (0.9,), 'submut': (0.1,), 'tournament': (8,), 'maxsize': (5000,)\} \\
              &                                                                                                \{'popsize': (10000,), 'functions': ('+\_-\_*\_aq\_plog\_sin\_cos',), 'linearscaling': (False,), 'sbrdo': (0.9,), 'submut': (0.1,), 'tournament': (8,), 'maxsize': (250,), 'sblibtype': ('p\_6\_9999',)\} \\
\bottomrule
\end{tabular}
\end{table}

\subsection{Additional Results}
\label{s:add_results}

\subsubsection{Subgroup analysis of black-box regression results}

\begin{figure}
    \centering
    \includegraphics[width=\textwidth]{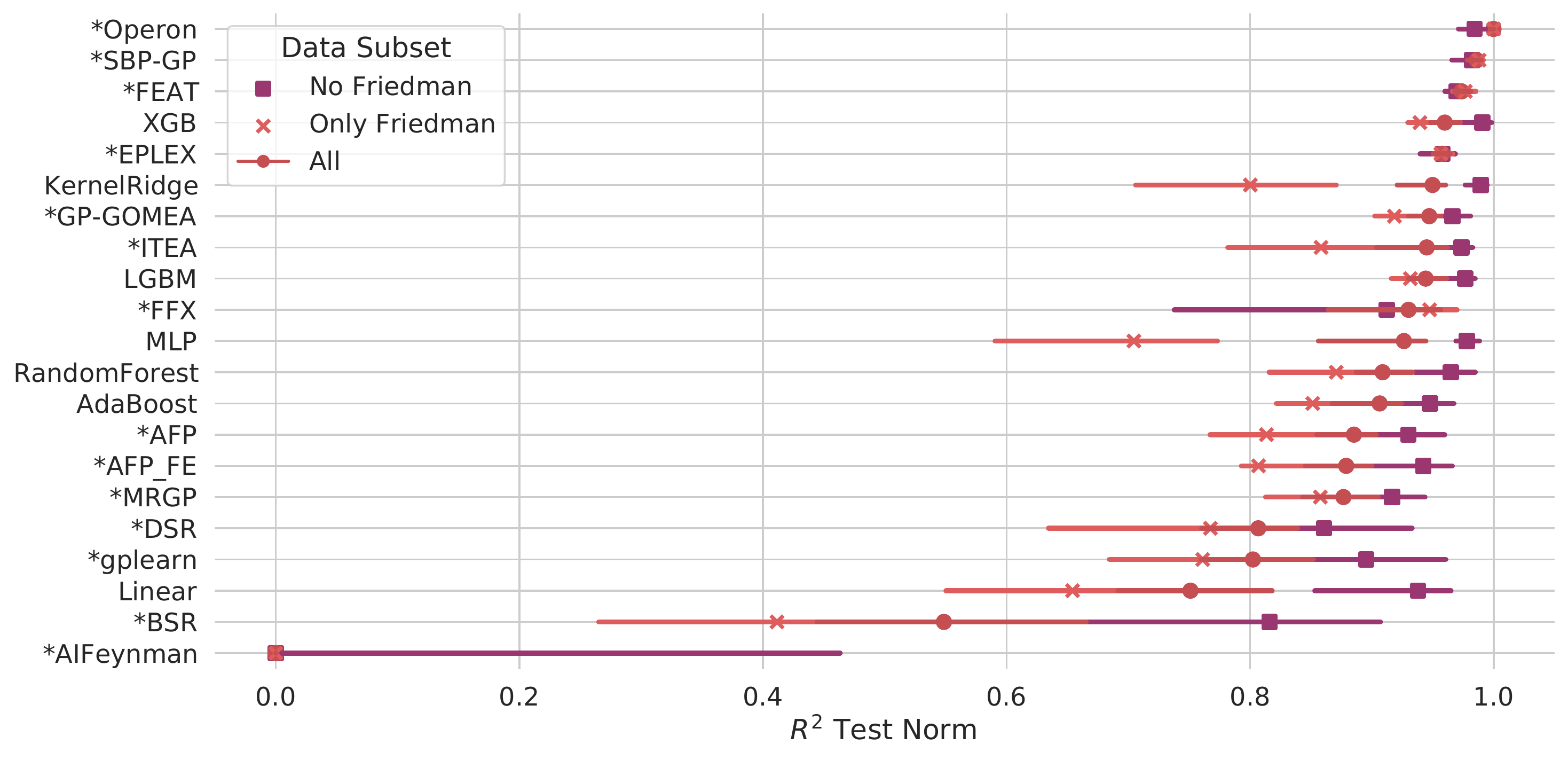}

    \caption{
        Comparison of normalized $R^2$ test scores on all black-box datasets, just the Friedman datatasets, and just the non-Friedman datasets.
    }
    \label{fig:friedman}
\end{figure}

Many of the black-box problems for regression in PMLB were originally sourced from \href{www.openml.org}{OpenML}. 
A few authors have noted that several of these datasets are sourced from~\citet{friedmanGreedyFunctionApproximation2001}'s synthetic benchmarks. 
These datasets are generated by non-linear functions that vary in degree of noise, variable interactions, variable importance, and degree of non-linearity. 
Due to their number, they may have an out-sized effect on results reporting in PMLB. 
In Fig.~\ref{fig:friedman}, we separate out results on this set of problems relative to the rest of PMLB. 
We do find that, relative to the rest of PMLB, the results on the Friedman datasets distinguish top-ranked methods more strongly than among the rest of the benchmark, on which performance between top-performing methods is more similar. 
In general, although we do see methods rankings change somewhat when looking at specific data groupings, we do not observe large differences. 
An exception is Kernel ridge regression, which performs poorly on the Friedman datasets but very well on the rest of PMLB.
We recommend that future revisions to PMLB expand the dataset collection to minimize the effect of any one source of data, and include subgroup analysis to identify which types of problems are best solved by specific methods. 

\begin{figure}
    \centering
    \includegraphics[width=\textwidth]{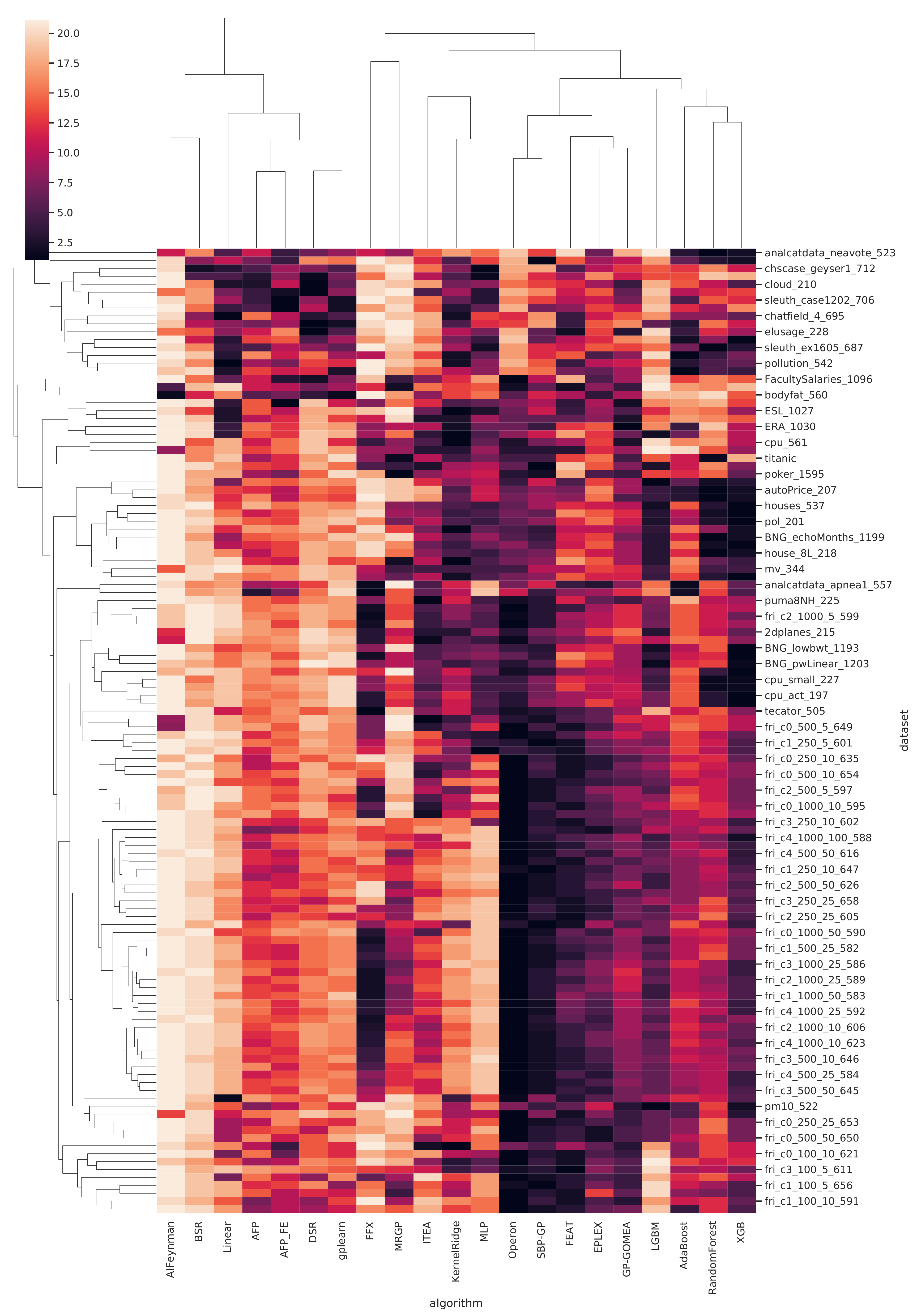}

    \caption{
        Rankings of methods by $R^2$ test score  on the black-box problems (lower is better). 
        Results are bi-clustered by SR method (columns) and dataset (rows). 
        Darker cells indicate that a method performs well on that dataset relative to its competitors.
        Note only a subset of the datasets are labelled due to space constraints.
    }
    \label{fig:bicluster}
\end{figure}

To get a better sense of the performance variability across methods and datasets, method rankings on each dataset are bi-clustered and visualized in Fig.~\ref{fig:bicluster}. 
Methods that perform most similarly across the benchmark are placed adjacent to each other, and likewise datasets that induce similar method rankings are grouped.
We note some expected groupings first: AFP and AFP\_FE, which differ only in fitness estimation, and FEAT and EPLEX, which use the same selection method, perform similarly. 
We also observe clustering among the Friedman datasets (names beginning with ``fri\_"), and again note stark differences between methods that perform well on these problems, e.g. Operon, SBP-GP, and FEAT, and those that do not, e.g. MLP. 
This view of the results also reveals a cluster of SR methods (AFP, AFP\_FE, DSR, gplearn) that perform well on a subset of real-world problems (analcatdata\_neavote\_523 - vineyard\_192) for which linear models also perform well.
Interestingly, for that problem subset, Operon's performance is mediocre relative to its strong performance on other datasets. 
We also note with surprise that DSR and gplearn exhibit performance similarity on par with AFP/AFP\_FE, and are the next most similar-performing methods (note the dendrogram connecting these columns). 

\subsubsection{Extended analysis of ground-truth regression results}
\begin{figure}
    \centering
    \includegraphics[width=0.6\textwidth]{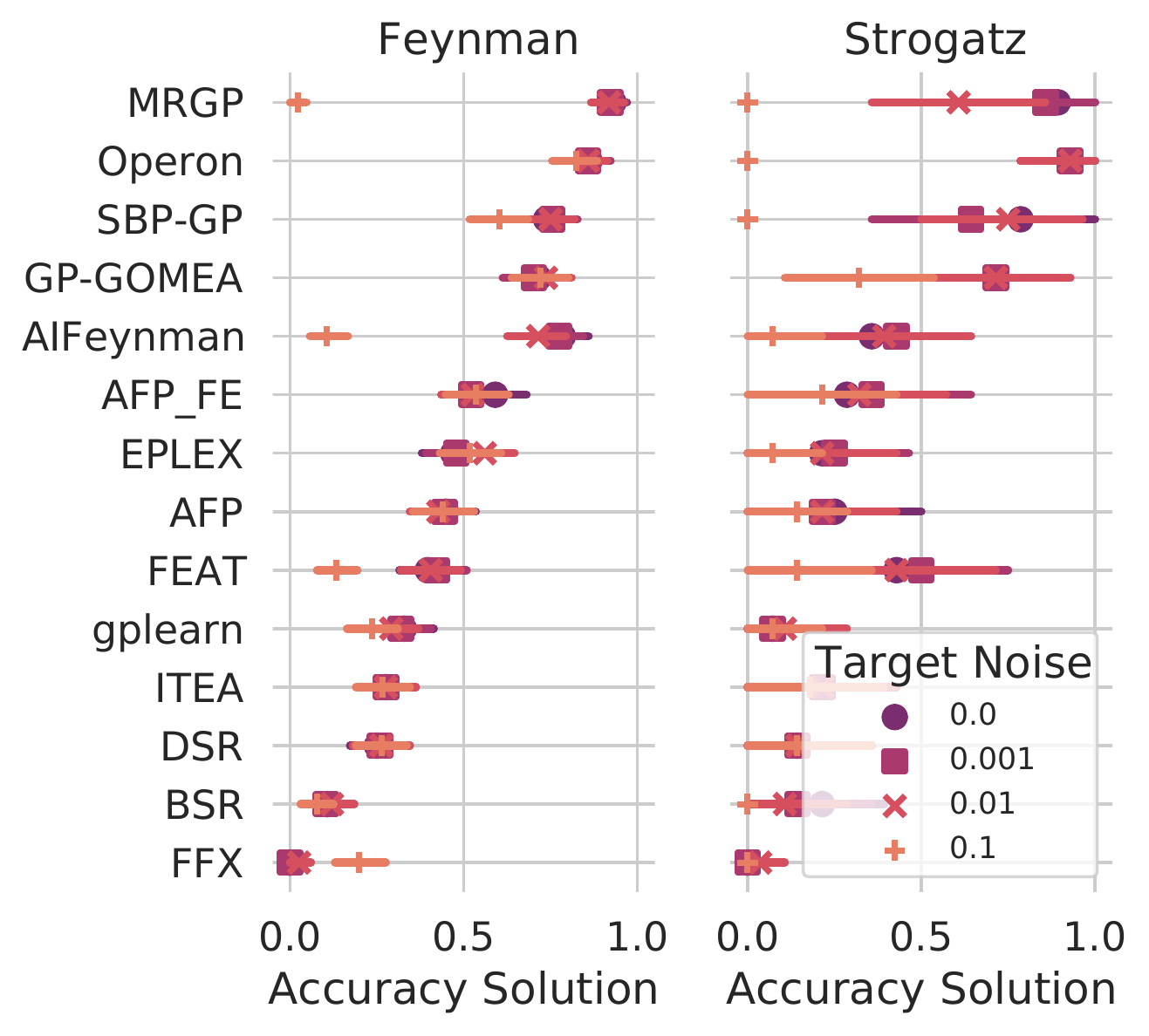}

    \caption{
        Subset comparison of "Accuracy Solutions", i.e. models with $R^2$>0.999 on the Feynman and Strogatz problems, differentiated by noise level.
    }
    \label{fig:acc_solns}
\end{figure}

% In Figs.~\ref{fig:feyn_strogatz} and \ref{fig:acc_solns}, we present the ground-truth results, subset by the data source. 
As noted in Sec.~\ref{s:disc}, despite Operon's good performance on black-box regression, it finds few models with symbolic equivalence. 
An alternative (and weaker) notion of solution is based on test set accuracy, which we show in Fig.~\ref{fig:acc_solns}; by this metric, the relative method performance corresponds more closely to that seen for black-box regression. 
We also note that methods that impose structural assumptions on the model (BSR, FEAT, ITEA, FFX) are worse at finding symbolic solutions, most of which do not match those assumptions (e.g. most processes in Table~\ref{tbl:strogatz}). 

% \begin{figure}
%     \includegraphics[width=\textwidth]{figs/results_sym_data/cat-pointplot-Symbolic-Solution-Rate-pct-by-Algorithm_Data-Group.pdf}
%     % \includegraphics[width=\textwidth]{figs/results_pmlb_r1/friedman_comparison_pairgrid-pointplot_r2_test.pdf}

%     \caption{
%         Subset comparison of solution rates on Feynman and Strogatz problems, differentiated by noise level.
%     }
%     \label{fig:feyn_strogatz}
% \end{figure}

\subsection{Statistical Tests}
\label{s:stats}

\begin{figure}
    \begin{minipage}{0.5\textwidth}
        \includegraphics[width=\textwidth]{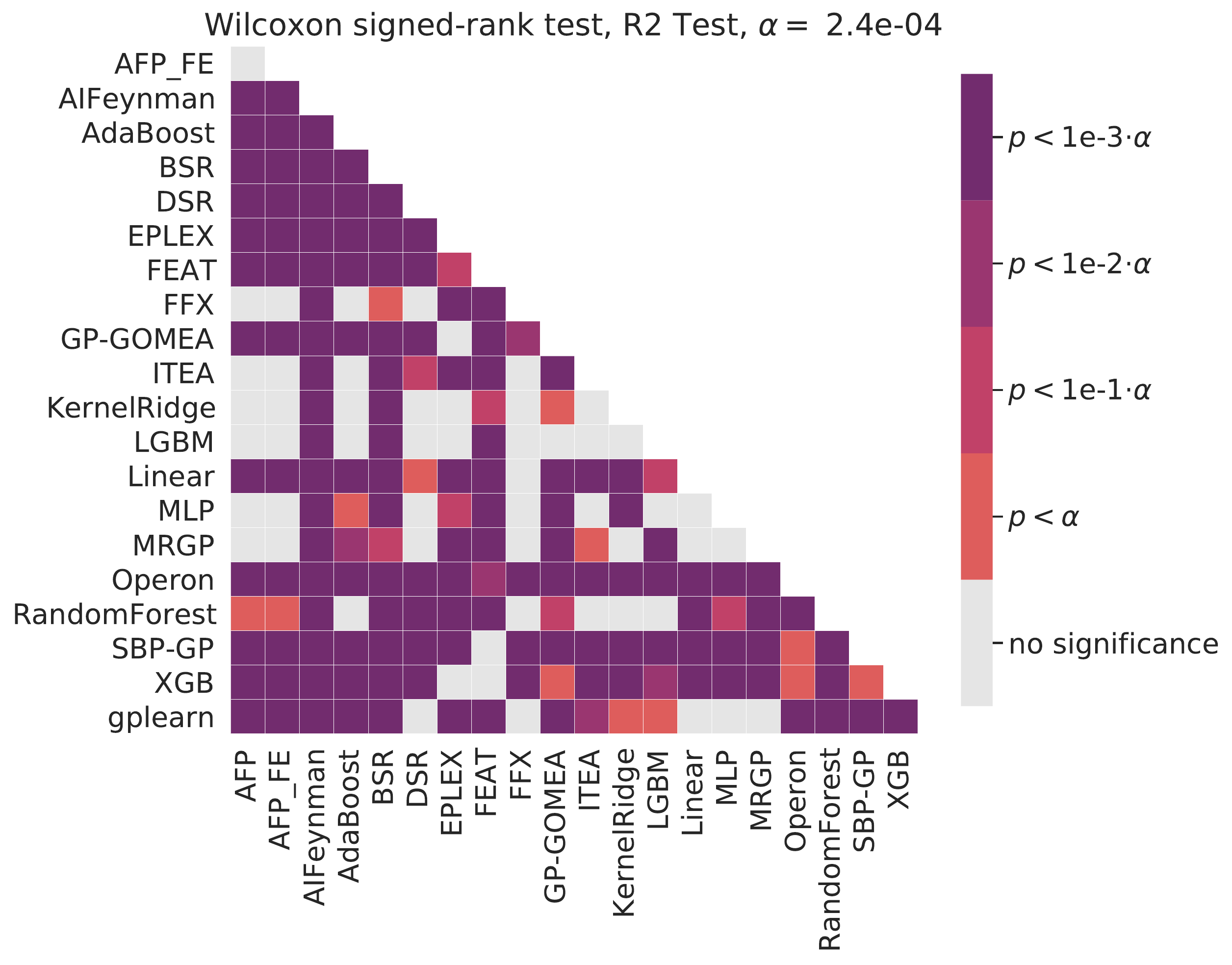}
    \end{minipage}
    % \hspace{0.02\textwidth}
    \begin{minipage}{0.5\textwidth}

        \includegraphics[width=\textwidth]{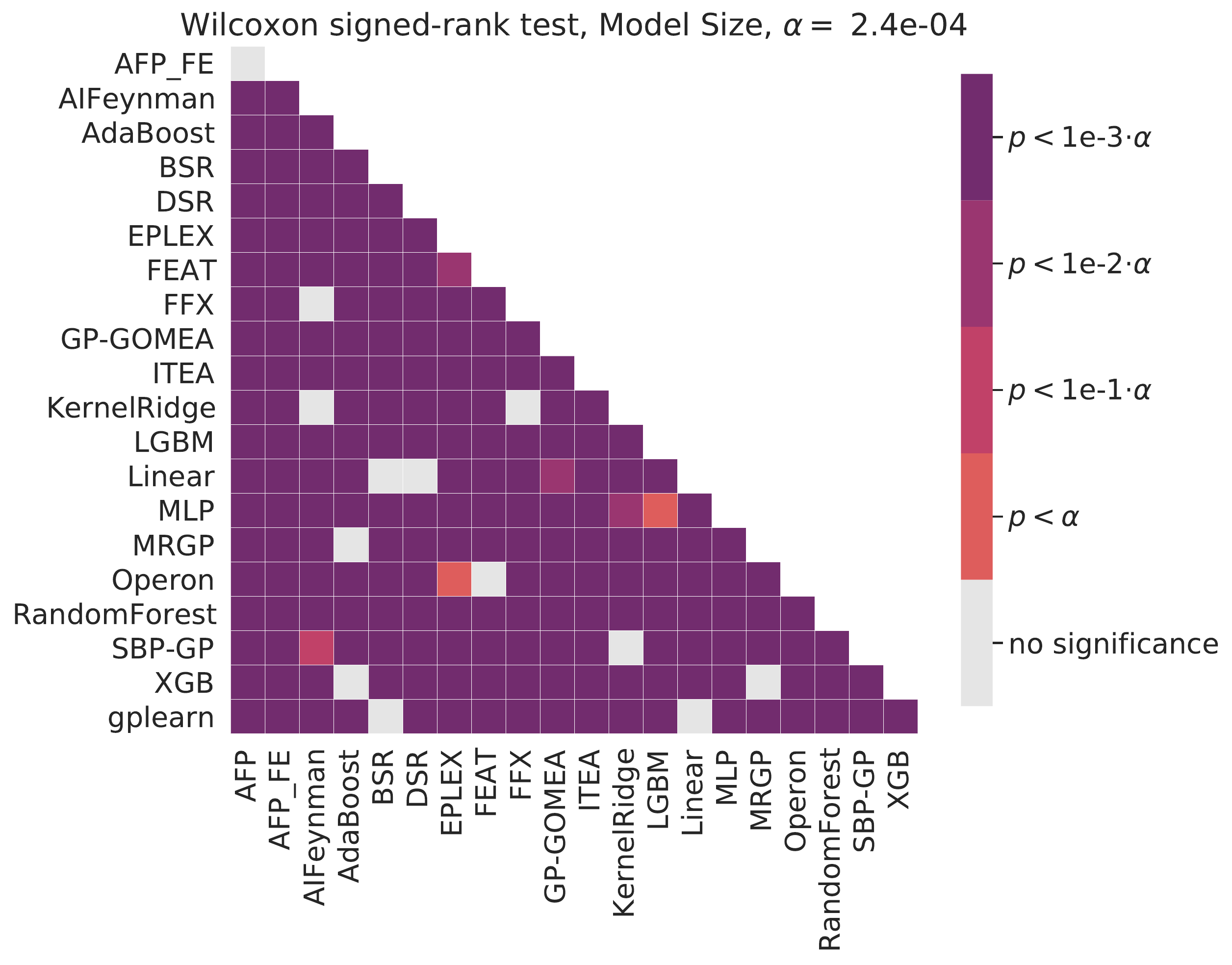}
    \end{minipage}
        \caption{ 
            Pairwise statistical comparisons on the black-box regression problems. 
            Wilcoxon signed-rank tests are used with a Bonferonni correction on $\alpha$ for multiple comparisons.
            (Left) $R^2$ test scores, (Right) model size. 
        }
        \label{fig:heat_stats_bb}
\end{figure}

\begin{figure}
    \begin{minipage}{0.5\textwidth}
        \includegraphics[width=\textwidth]{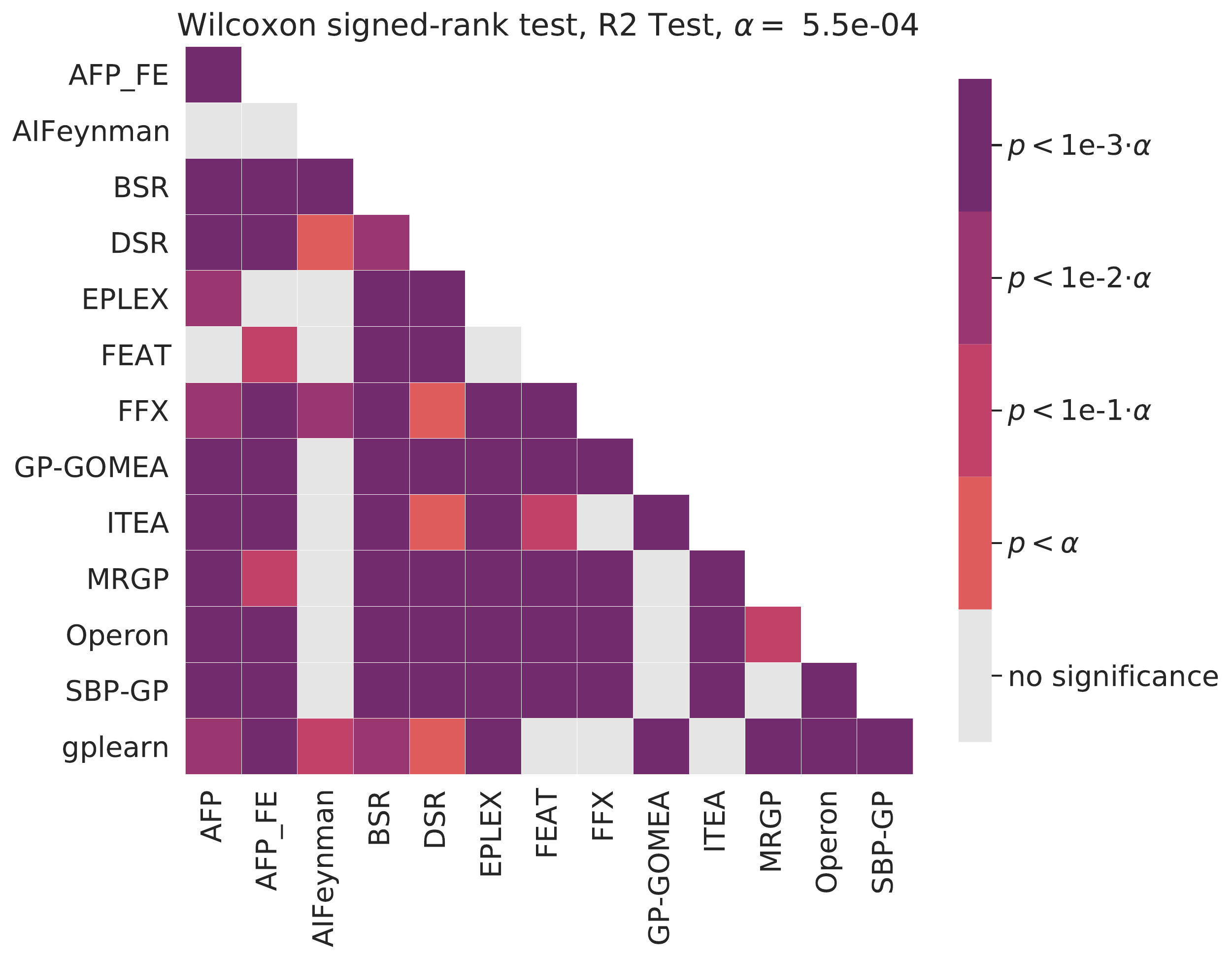}
    \end{minipage}
    % \hspace{0.02\textwidth}
    \begin{minipage}{0.5\textwidth}
        \includegraphics[width=\textwidth]{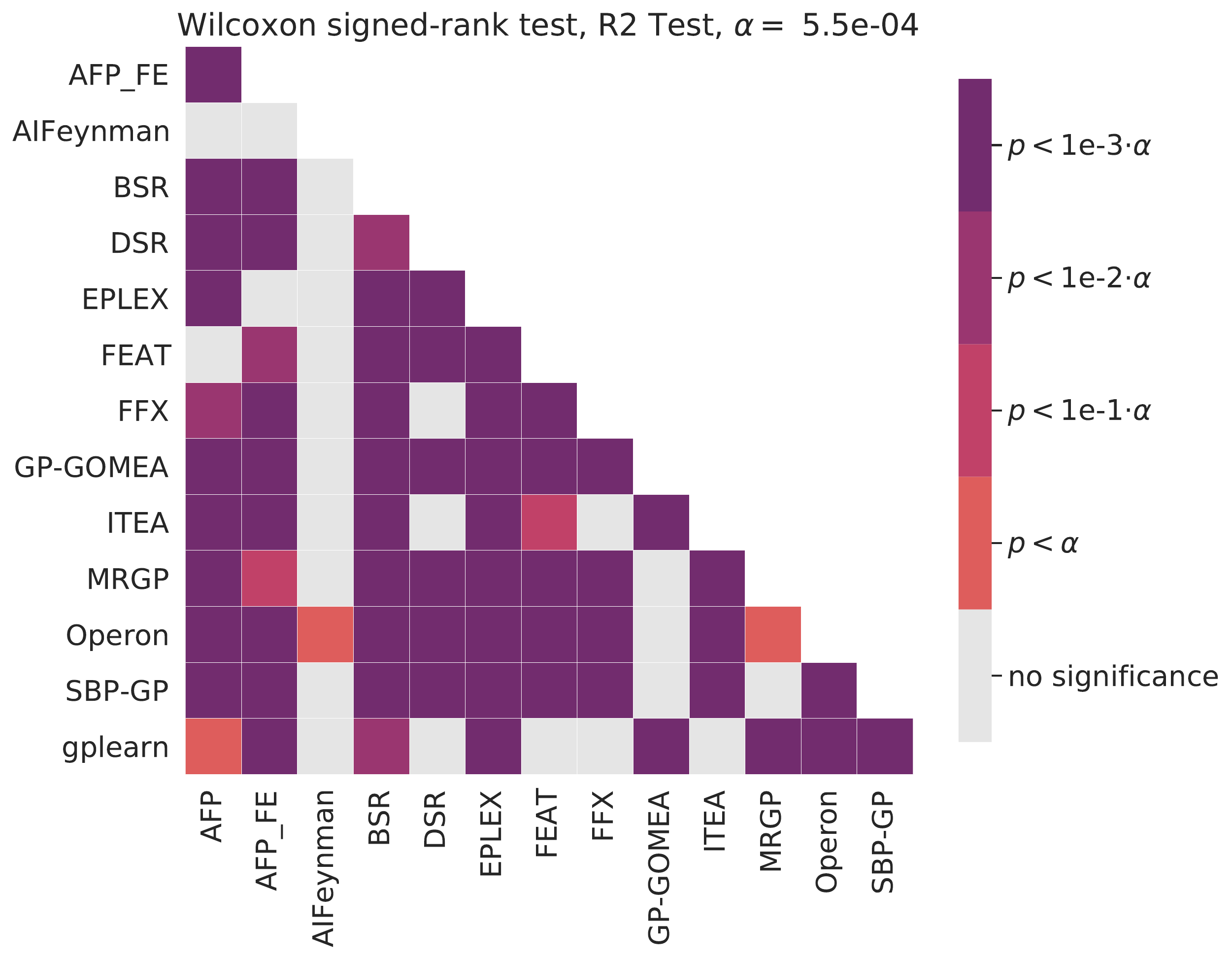}
    \end{minipage}
    \caption{ 
        Pairwise statistical comparisons of $R^2$ test scores on the ground-truth regression problems. 
        We report Wilcoxon signed-rank tests with a Bonferonni correction on $\alpha$ for multiple comparisons.
        (Left) target noise of 0, (Right) target noise of 0.01. 
    }
    \label{fig:heat_stats_r2_sr}
\end{figure}

\begin{figure}
    \begin{minipage}{0.5\textwidth}
        \includegraphics[width=\textwidth]{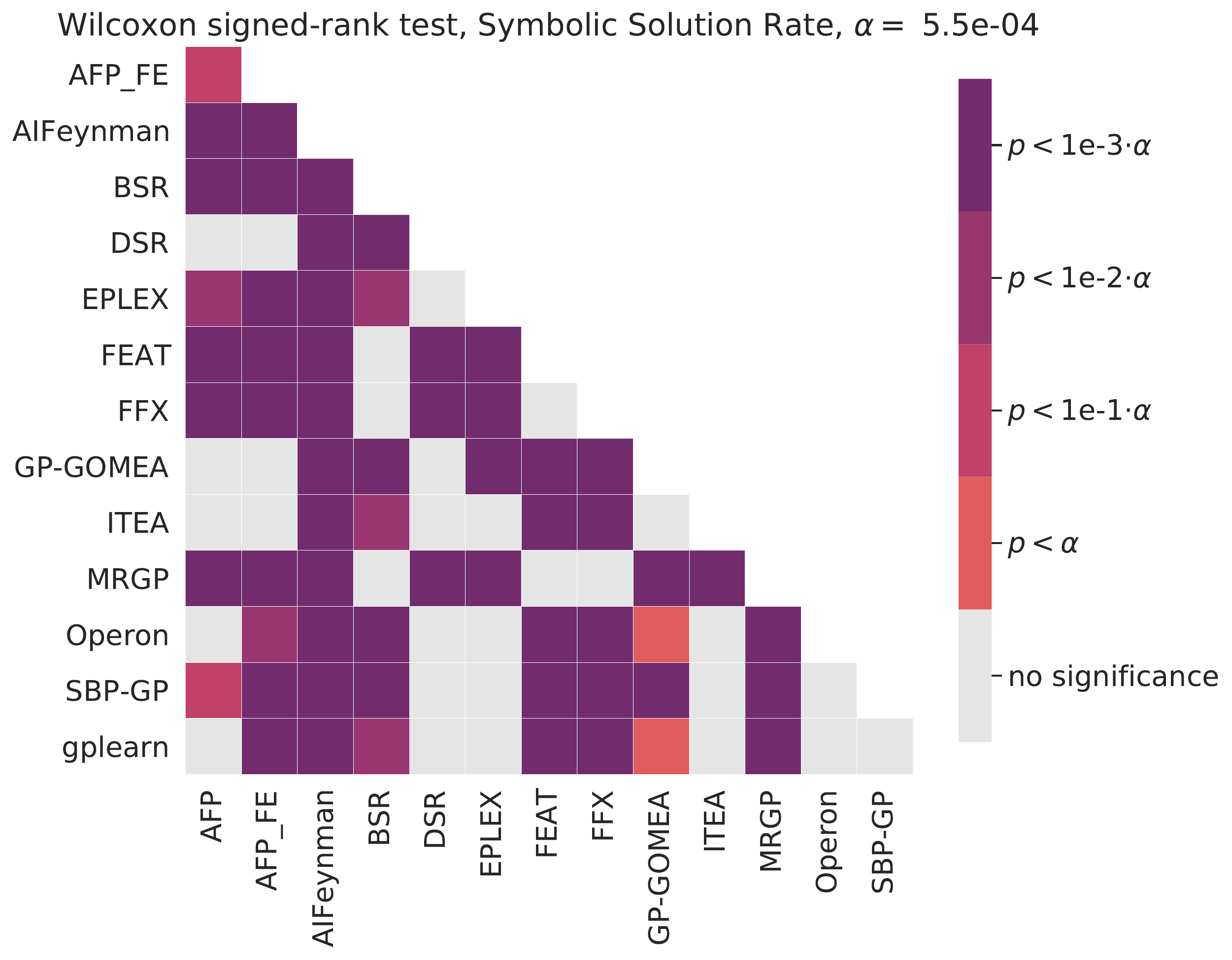}
    \end{minipage}
    \begin{minipage}{0.5\textwidth}
        \includegraphics[width=\textwidth]{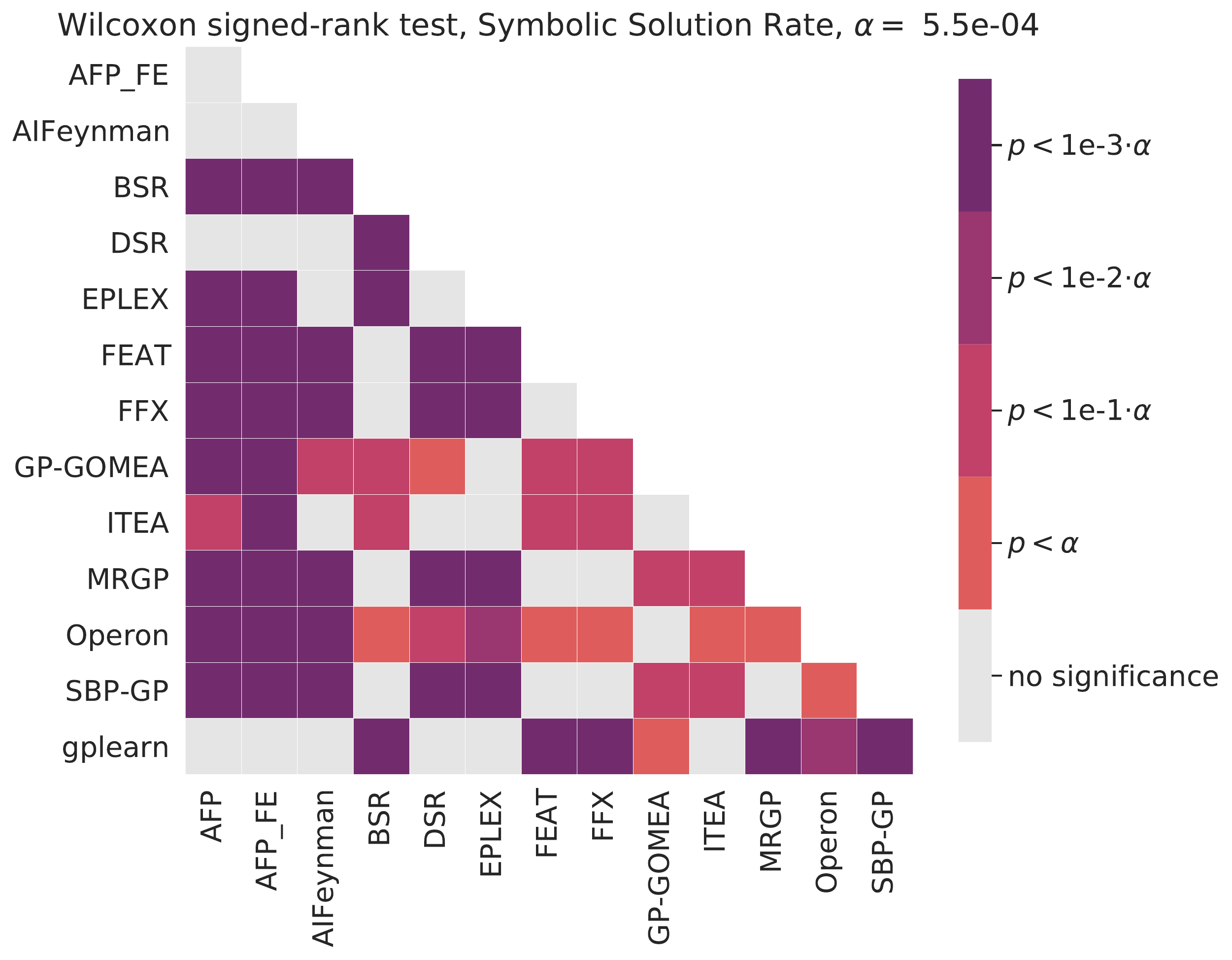}
    \end{minipage}
    \caption{ 
        Pairwise statistical comparisons of solution rates on the ground-truth regression problems. 
        We report Wilcoxon signed-rank tests with a Bonferonni correction on $\alpha$ for multiple comparisons.
        (Left) target noise of 0, (Right) target noise of 0.01. 
    }
    \label{fig:heat_stats_soln_sr}
\end{figure}

Figures~\ref{fig:heat_stats_bb}-\ref{fig:heat_stats_soln_sr} give summary significance levels of pairwise tests of significance between estimators on the black-box and ground-truth problems. 
All pair-wise statistical tests are Wilcoxon signed-rank tests. 
A Bonferroni correction was applied, yielding the $\alpha$ levels given in each. 
This methodology for assessing statistical significance is based on the recommendations of~\citet{demsarStatisticalComparisonsClassifiers2006a} for comparing multiple estimators over many datasets.
These figures are intended to complement Figures~\ref{fig:pmlb_perf}-\ref{fig:symbolic_solns} in which effect sizes are shown. 

\end{document}